\documentclass[11pt]{article}
\usepackage{lineno}
\usepackage{wrapfig}
\usepackage{fullpage}
\usepackage{natbib}

\usepackage{microtype}
\usepackage{graphicx}
\usepackage{float}
\usepackage{comment}
\usepackage{booktabs}
\usepackage[hidelinks]{hyperref} 
\usepackage{amsmath, amssymb, mathtools, amsthm}
\usepackage{cleveref} 
\usepackage[a4paper, margin=1in]{geometry}

\theoremstyle{plain}
\newtheorem{example}{Example}
\newtheorem{theorem}{Theorem}[section]

\newtheorem{lemma}[theorem]{Lemma}

\theoremstyle{definition}

\newtheorem{assumption}[theorem]{Assumption}
\theoremstyle{remark}
\newtheorem{remark}[theorem]{Remark}

\renewcommand{\section}{%
  \@startsection{section}{1}{\z@}%
                {-2.0ex \@plus -0.5ex \@minus -0.2ex}%
                { 1.5ex \@plus  0.3ex \@minus  0.2ex}%
                {\large\bf\raggedright}%
}

\renewcommand{\subsection}{%
  \@startsection{subsection}{2}{\z@}%
                {-1.8ex \@plus -0.5ex \@minus -0.2ex}%
                { 0.8ex \@plus  0.2ex}%
                {\normalsize\bf\raggedright}%
}

\renewcommand{\subsubsection}{%
  \@startsection{subsubsection}{3}{\z@}%
                {-1.5ex \@plus -0.5ex \@minus -0.2ex}%
                { 0.5ex \@plus  0.2ex}%
                {\normalsize\bf\raggedright}%
}

\makeatletter
\renewcommand{\section}{%
  \@startsection{section}{1}{\z@}%
                {-2.0ex \@plus -0.5ex \@minus -0.2ex}%
                { 1.5ex \@plus  0.3ex \@minus  0.2ex}%
                {\large\bf\raggedright}%
}

\renewcommand{\subsection}{%
  \@startsection{subsection}{2}{\z@}%
                {-1.8ex \@plus -0.5ex \@minus -0.2ex}%
                { 0.8ex \@plus  0.2ex}%
                {\normalsize\bf\raggedright}%
}

\renewcommand{\subsubsection}{%
  \@startsection{subsubsection}{3}{\z@}%
                {-1.5ex \@plus -0.5ex \@minus -0.2ex}%
                { 0.5ex \@plus  0.2ex}%
                {\normalsize\bf\raggedright}%
}
\makeatother

\usepackage[textsize=tiny]{todonotes} 
\usepackage{url}            
\usepackage{xspace}
\usepackage{multirow}
\usepackage{colortbl}
\usepackage{enumerate}

\def\R{\mathbb{R}}

\newcommand{\bfx}{\mathbf{x}}
\newcommand{\bft}{\boldsymbol{\theta}}

\newcommand{\bfy}{\mathbf{y}}

\newcommand{\li}{L} 

\newcommand{\bfu}{\mathbf{u}}

\newcommand{\cld}{\mathcal{D}}

\renewcommand{\comment}[1]{}

\newcommand{\la}{\left\langle}
\newcommand{\ra}{\right\rangle}
\newcommand{\lnr}{\left\|}
\newcommand{\rnr}{\right\|}

\newcommand{\blr}{\texttt{BLUR}\xspace}
\definecolor{lightyellow}{RGB}{255, 255, 0}  
\definecolor{bluegreen}{RGB}{0,150,200} 

\usepackage{caption}

\title{BLUR: A Bi-Level Optimization Approach for LLM Unlearning}
\author{Hadi Reisizadeh\textsuperscript{1,$\ast$}, Jinghan Jia\textsuperscript{2,$\ast$}, Zhiqi Bu$^6$, Bhanukiran Vinzamuri$^3$, \\
Anil Ramakrishna$^6$, Kai-Wei Chang$^{3,4}$, Volkan Cevher$^{3,5}$, \\
Sijia Liu$^2$, Mingyi Hong$^{1,3}$ \\ 
\\
$^1$ University of Minnesota, $^2$ Michigan State University, \\ 
$^3$ Amazon AGI, $^4$ UCLA, $^5$ LIONS EPFL, 
$^6$ Meta \\
\texttt{hadir@umn.edu} 
}
\date{} 

\begin{document}

\maketitle  
\begingroup
    \renewcommand\thefootnote{}\footnotetext{$\ast$ Equal contribution.}
    \endgroup 
\begin{abstract}
Enabling large language models (LLMs) to unlearn knowledge and capabilities acquired during training has proven vital for ensuring compliance with data regulations and promoting ethical practices in generative AI. Although there are growing interests in developing various unlearning algorithms, it remains unclear how to best formulate the unlearning problem. The most popular formulation uses a weighted sum of forget and retain loss, but it often leads to performance degradation due to the inherent trade-off between forget and retain losses. In this work, we argue that it is important to model the hierarchical structure of the unlearning problem, where the forget problem (which \textit{unlearns} certain knowledge and/or capabilities) takes priority over the retain problem (which preserves model utility). This hierarchical structure naturally leads to a bi-level optimization formulation where the lower-level objective focuses on minimizing the forget loss, while the upper-level objective aims to maintain the model's utility. Based on this new formulation, we propose a novel algorithm, termed Bi-Level UnleaRning (\blr), which not only possesses strong theoretical guarantees but more importantly, delivers superior performance. In particular, our extensive experiments demonstrate that~\blr consistently outperforms all the state-of-the-art algorithms across various unlearning tasks, models, and metrics. Codes are available at \url{https://github.com/OptimAI-Lab/BLURLLMUnlearning}.
\end{abstract} 

\vspace{-0.2cm}
\section{Introduction}\label{sec:intro}
\vspace{-0.2cm}
Large language models (LLMs) have illustrated exceptional power in text generation that closely mimics human interactions~\cite{touvron2023llama}. However, these models are trained and fine-tuned on large datasets that are usually collected from the web. This raises ethical and privacy issues, such as generating biased~\cite{kotek2023gender,motoki2023more}, toxic, private, illegal responses~\cite{nasr2023scalable,wen2023unveiling,karamolegkou2023copyright,sun2024trustllm}, and potential guides on developing bioweapons and cyberattacks~\cite{barrett2023identifying,li2024wmdp}. LLM unlearning has emerged as a useful technique to mitigate these concerns, which \textit{forget} these toxic data influences from the pre-trained LLMs and ensure the unlearned models are safe for various applications while preserving the model's overall utility after the unlearning phase.
\vspace{-0.2cm}
\subsection{Challenges and Our Contributions}
\vspace{-0.2cm}
 Unlearning in LLMs introduces unique challenges due to the massive size and complexity of their training datasets, as well as the risks of memorizing biases, sensitive information, and harmful content. Another challenge is to precisely define the \textit{unlearning targets}, such as sample data points in the training set or knowledge concepts that must be forgotten during the unlearning phase, which usually leads to task-based solutions~\cite{jang2022knowledge,ilharco2022editing,eldan2023whos}. Also, reliable evaluation mechanisms for LLM unlearning is still lacking, and it is shown that sensitive information can be retrieved by reverse engineering techniques such as relearning~\cite{hu2024jogging,lynch2024eight} and jailbreaking attacks~\cite{lucki2024adversarial,shumailov2024ununlearning}.  
 
\noindent{\bf How to Balance \textit{Forget} and \textit{Retain}?} One {\it key algorithmic challenge} that we attempt to address in this work is that during the unlearning process, how to best {\it balance} between the task of  `unlearning' knowledge/capabilities, and `retaining' model utility.  Indeed, it has been generally observed that removing undesired information can degrade the model's utility, while insufficient forgetting may fail to achieve unlearning goals. Therefore, it is critical to ensure that these two tasks are carefully solved together to ensure that {\it both} are achieved eventually. However, in almost all existing works~\cite{liu2022continual,yao2023large,maini2024tofu,eldan2023whos,zhang2024negative}, the problem of LLM unlearning has been formulated as a {\it weighted sum} of the forget and retain losses, with a {\it fixed} weighting factor used to indicate the relative importance of the two tasks. Despite the simplicity of this formulation, it fails to fully capture the dynamic nature of unlearning, that is, the importance of the retain and forget loss often changes as the optimization goes, leading to model performance degradation. See Sec. \ref{sub:bilevel} for more detailed discussion on this point.

\noindent{\bf How to Model the Hierarchical Structure?} Beyond algorithmic considerations, a more fundamental challenge lies in defining an effective formulation for the unlearning problem. Should we aim to \textit{forget} and \textit{retain} simultaneously? In many cases, the answer is no. Unlearning is typically necessitated in scenarios where the removal of certain sensitive information is critical. This may be driven by ethical and legal compliance requirements, such as adhering to privacy regulations (e.g., GDPR~\cite{gdpr2016general}), or by the need to address fairness concerns by mitigating biases in the model \cite{binns2018fairness}. In such cases, failure to completely remove the identified information is unacceptable, making the \textit{forget} task a priority. Once the \textit{forget} task is completed successfully, the remaining capacity of the model if sufficient can then be leveraged to focus on the \textit{retain} task~\cite{liu2024towards}. This hierarchical prioritization ensures that sensitive information is effectively removed while still striving to preserve the model’s utility for its intended applications. Unfortunately, none of the existing unlearning works have considered this key aspect. 

\subsection{Our contributions} 
This work proposes to approach the unlearning problem from a fresh perspective. Instead of treating unlearning as a binary process of simply forgetting specific information while retaining the rest, we argue that we should prioritize and structure these tasks hierarchically. Specifically, the forget task should take precedence to ensure that sensitive or harmful information is thoroughly removed before addressing the retain task. This perspective allows for a more principled approach to unlearning. Interestingly, the algorithm derived from the new perspective {\it dynamically} adjusts its emphasis on the forget and retain loss during the optimization process, addressing the previously mentioned `balancing' challenges. More concretely, our contributions are listed below:

\begin{wrapfigure}{r}{0.5\textwidth}
\vspace{-0.5cm}
   \includegraphics[width=\linewidth]{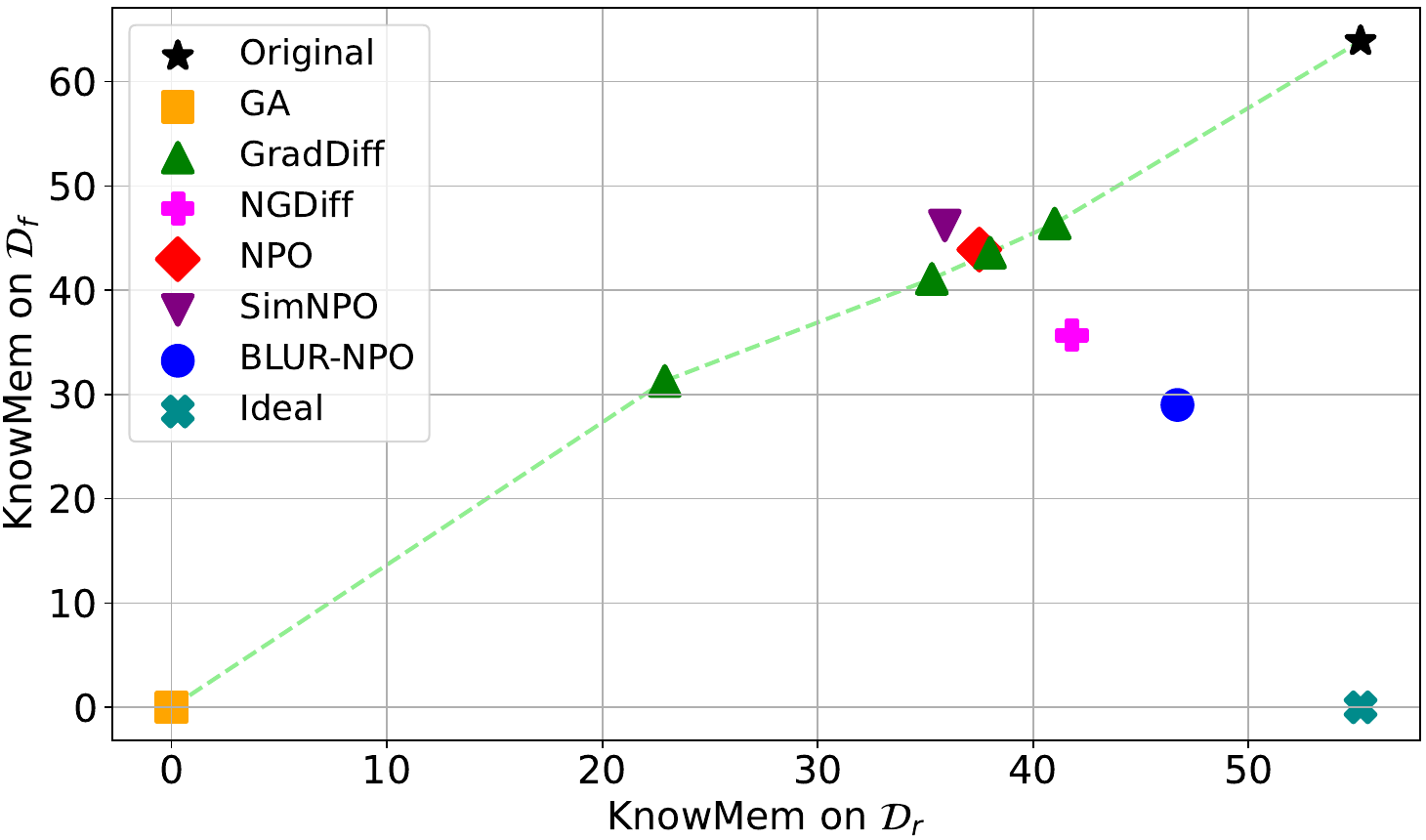}
    \caption{Trade-off between Knowledge memorization values on the forget set (vertical axis, the lower the better) and retain datasets (horizontal axis, the higher the better) using different unlearning methods. Training is done using LLaMA2-7B model, evaluated using the MUSE-News dataset. We run GradDiff with various values of the regularization term $\lambda$, as defined in~\eqref{eq:reg_f}. }
    \label{fig:trade-off-news}
    \vspace{-0.5cm}
    \end{wrapfigure}

    \noindent {\bf (1).} Observing that the aforementioned hierarchical structure is critical for understanding the unlearning problem, we begin by formulating it as a bi-level optimization problem, where the {\it lower-level} problem focuses on identifying a set of solutions that minimize the forget loss, ensuring that sensitive or undesirable information is effectively removed. The {\it upper-level} problem then selects one of these solutions from the lower-level that minimizes the retrain loss, thereby preserving as much of the remaining knowledge as possible. This formulation introduces a principled way to balance these competing objectives, leading to a novel and flexible approach to unlearning.
   
    \vspace{-0.1cm}
    \noindent {\bf (2).} We develop a novel algorithm named Bi-Level UnleaRning  (\blr), which solves the above bi-level unlearning problem. At a high level, this algorithm takes a gradient descent step of the forget objective and then updates the retain loss only in the direction that is {\it orthogonal} to the computed forget gradient. Such an orthogonalization is achieved by carefully and dynamically updating the weights that balance the forget and retain loss when updating the LLM parameters. In addition, we show that the algorithm converges to certain desired solutions. 

    \noindent {\bf (3).} We conduct several experiments on widely used datasets, including MUSE~\cite{shi2024muse} and WMDP~\cite{li2024wmdp}. We demonstrate the effectiveness of \blr across diverse unlearning benchmarks and evaluation metrics, demonstrating that ~\blr outperforms a number of state-of-the-art LLM unlearning algorithms; see, e.g.,  \textbf{Fig~\ref{fig:trade-off-news}}. In particular, our algorithm~\blr on the MUSE-News dataset achieves $19\%$ and $11.7\%$ higher performance than the state-of-the-art baselines in unlearning efficiency and model utility, respectively, on the MUSE-News dataset. Surprisingly, by adapting and~\blr using the loss functions designed in RMU, the algorithm outperforms RMUs, achieving $16.6\%$ higher performance in unlearning efficiency while maintaining the same level of model utility.  

\vspace{-0.2cm}
\subsection{Other Related Works} 
\vspace{-0.2cm}
\noindent\textbf{Machine Unlearning for Non-LLMs.}
The concept of machine unlearning (MU) originated from data protection regulations such as the \textit{right to be forgotten}~\cite{rosen2011right}. The MU has emerged in various applications such as image classification~\cite{sekhari2021remember,fan2023salun}, text-to-image generation~\cite{gandikota2023erasing,zhang2024defensive}, federated learning~\cite{liu2020federated,liu2023survey,che2023fast}, graph neural networks~\cite{chen2022graph,wu2023certified}, and recommendation~\cite{sachdeva2024machine,xu2023netflix}.

\noindent\textbf{LLM Unlearning.}
Retraining LLMs from scratch is often infeasible due to the amount of training datasets. Hence, removing undesirable information from the pre-trained model is critical for practical unlearning. Although solving the LLM unlearning problem with an initial pre-trained model appears easy, the challenges of choosing suitable losses, especially forget loss introduce new complexities in achieving the optimal balance between unlearning and utility. Some works~\cite{thudi2022unrolling,yao2023large,maini2024tofu} have utilized the gradient ascent (GA) approach on the prediction loss over the undesirable dataset (forget set). Even though this approach is intuitive, as it implies reversing gradient descent, the performance of gradient ascent-based approaches remains unsatisfactory, particularly in terms of model utility due to the unboundedness of gradient ascent loss. To address this issue, efficient forget losses are developed, such as preference optimization (PO)~\cite{rafailov2023direct}, negative preference optimization (NPO)~\cite{zhang2024negative}, and simple negative preference optimization (SimNPO)~\cite{fan2024simplicity}. During LLM preference alignment, PO replaces true information with random information for the forget set, while NPO treats the forget set as negative samples. SimNPO eliminates the dependency of the forget loss on the reference model and provides an improved version of NPO. Recently, another related work~\cite{bu2024unlearning} studies LLM unlearning as a regularized multi-task optimization problem, where one task optimizes the forget loss objective and the other preserves model utility. A normalized gradient difference method, termed NGDiff, is then developed based on dynamic scalarization. All these approaches do not address the hierarchical structure of the unlearning problem.
\vspace{-0.2cm}
\section{LLM Unlearning as a Bi-level Optimization Problem}\label{sec:main}
\vspace{-0.2cm}
\subsection{Preliminaries}
\vspace{-0.2cm}
We start by defining some notations. Let  $\cld_f$ be the forget dataset that contains the data whose influence on the model is to be removed, and $\cld_r$ as a retain dataset, which includes samples that help preserve the model's utility. The LLM unlearning is typically modeled in the following manner~\cite{liu2022continual,yao2023large,maini2024tofu,eldan2023whos,zhang2024negative}
\begin{align}\label{eq:reg_f}
    \!\min_{\bft}  \mathbb{E}_{(x, y) \in \mathcal{D}_f} \!\big[\ell_f(y \!\!\mid\! x; \bft)\big]\!+\! \lambda \mathbb{E}_{(x,y) \in \mathcal{D}_r} \!\big[\ell_r(y \!\mid\! x; \bft)\big],
\end{align}
where $\ell_f(y \!\mid\! x; \bft)$, $\ell_r(y \!\mid\! x; \bft)$ represent the forget and retain prediction loss, respectively, computed using the model parameter $\bft$ for an input $x$ with respect to the response $y$. Here, the parameter $\lambda\geq 0$ is a regularization term used to balance forget and retain.  The retain loss is typically cross-entropy loss, given by
\begin{align}\label{eq:cross}
    \ell_r(y\mid x;\bft) = -\log(\pi(y \mid x;\bft)),
\end{align}
where $\pi(y \mid x;\bft)$ is the output probability distribution of the current model $\bft$.  Commonly used forget losses are given below: \\
$\bullet \ \ell_{\text{GA}}$~\cite{maini2024tofu,thudi2022unrolling} represents the gradient descent technique on the negative prediction loss leading the updated model's predictions to diverge from the pre-trined model's. This loss function is defined as
    \begin{align}\label{eq:l_ga}
        \ell_{\text{GA}} = \log(\pi(y \mid x;\bft)).
    \end{align}
$\bullet \ \ell_f = \ell_{\text{NPO},\beta}$ for a given $\beta\geq 0$~\cite{zhang2024negative} treats the forget set as negative examples, given by
\begin{align}\label{eq:npo}
\!\ell_{\text{NPO}, \beta}(y \!\mid \!x;\bft) \!=\! \frac{2}{\beta} 
\log \!\left(\!1 \!+\! \left(\frac{\pi(y \!\mid\! x;\bft)}{\pi(y \!\mid\! x;\bft_0)}\!\right)^\beta\right),
\end{align}
where $\pi(y \mid x;\bft_0)$ represents the reference probability distribution of the pre-trained model $\bft_0$. 
$\bullet \ \ell_f = \ell_{\text{SimNPO},\beta,\alpha}$ for given $\beta,\alpha\geq 0$~\cite{fan2024simplicity} adopts a reference-free reward formulation that is normalized by sequence length, defined as
    \begin{align}\label{eq:simnpo}
        &\ell_{\text{SimNPO},\beta,\alpha}(y \!\mid \!x;\bft) = -\frac{2}{\beta} \log \sigma \left( -\frac{\beta}{|y|} \log \pi (y \!\mid \!x;\bft) \!-\! \alpha \right),
        \end{align}
        where $|y|$ is the response length and $\alpha \geq 0$ serves as the reward margin parameter. For simplicity, let $f(\bft)\!:=\!\mathbb{E}_{(x, y) \in \mathcal{D}_f} \!\big[\ell_f(y \!\!\mid\! x; \bft)\big],$ and $r(\bft)\!:=\!\mathbb{E}_{(x, y) \in \mathcal{D}_r}\!\big[\ell_r(y \!\mid\! x; \bft)\big]$. 

Solving the regularized optimization problem in~\eqref{eq:reg_f} using the gradient descent technique, we obtain the update direction and the corresponding update scheme given by

\begin{minipage}{0.5\textwidth}
\begin{align}
    \hat{u}(\bft) &= \nabla f(\bft) + \lambda \nabla r(\bft), \label{eq:u_hat}
\end{align}
\end{minipage}
\hfill
\begin{minipage}{0.5\textwidth}
\begin{align}
    \hat{\bft}(t+1) &= \hat{\bft}(t) - \eta \hat{u}(\hat{\bft}(t)). \label{eq:theta_hat}
\end{align}
\end{minipage}
\begin{wraptable}{r}{0.5\textwidth}
\caption{Summary of unlearning methods with their retain and forget losses.}
\vspace{-0.2cm}
\centering
\renewcommand{\arraystretch}{1.0} 
\setlength{\tabcolsep}{5pt}
\begin{tabular}{@{}>{\small \centering\arraybackslash}p{0.23\columnwidth}>{\centering\arraybackslash}p{0.12\columnwidth}>{\centering\arraybackslash}p{0.12\columnwidth}@{}}
\toprule
\textbf{Unlearning Method} & \textbf{Retain Loss} & \textbf{Forget Loss} \\ \midrule
GA~\cite{maini2024tofu} & N/A & \eqref{eq:l_ga} \\ 
GradDiff~\cite{liu2022continual} & \eqref{eq:cross} & \eqref{eq:l_ga} \\ 
NPO~\cite{zhang2024negative} & \eqref{eq:cross} & \eqref{eq:npo} \\ 
SimNPO~\cite{fan2024simplicity} & \eqref{eq:cross} & \eqref{eq:simnpo} \\  
\bottomrule
\end{tabular}
\label{tab:unlearning_methods}
\vspace{-0.5cm}
\end{wraptable}

where $\eta>0$ is the learning rate. Utilizing the retain loss and forget objective in the gradient direction defined in~\eqref{eq:u_hat}, we derive various unlearning methods, summarized in 
\textbf{Table~\ref{tab:unlearning_methods}}.
\vspace{-0.2cm}
\subsection{Unlearning as Bi-level Optimization}\label{sub:bilevel}
\vspace{-0.2cm}
As mentioned in the introduction, one key challenge in formulations \eqref{eq:reg_f} is that the weighted sum of two losses cannot properly prioritize one task (e.g. \textit{forget}) over the other (e.g., \textit{retain}). To illustrate this point, we consider a set of simple numerical experiments where we measure how the update direction $\hat{u}(\bft)$ aligns with both forget and retain gradient functions. To this end, let us define the normalized alignment as:
\begin{align}\label{eq:alig}
    \! A_f(\bft)\!:=\!\frac{\la \nabla f(\bft),\hat{u}(\bft)\ra}{\|\nabla f(\bft)\|^2}, \! \quad \!\! A_r(\bft)\!:=\!\frac{\la \nabla r(\bft),\hat{u}(\bft)\ra}{\|\nabla r(\bft)\|^2}. 
\end{align}
\begin{wrapfigure}{r}{0.5\textwidth}
    \includegraphics[width=\linewidth]{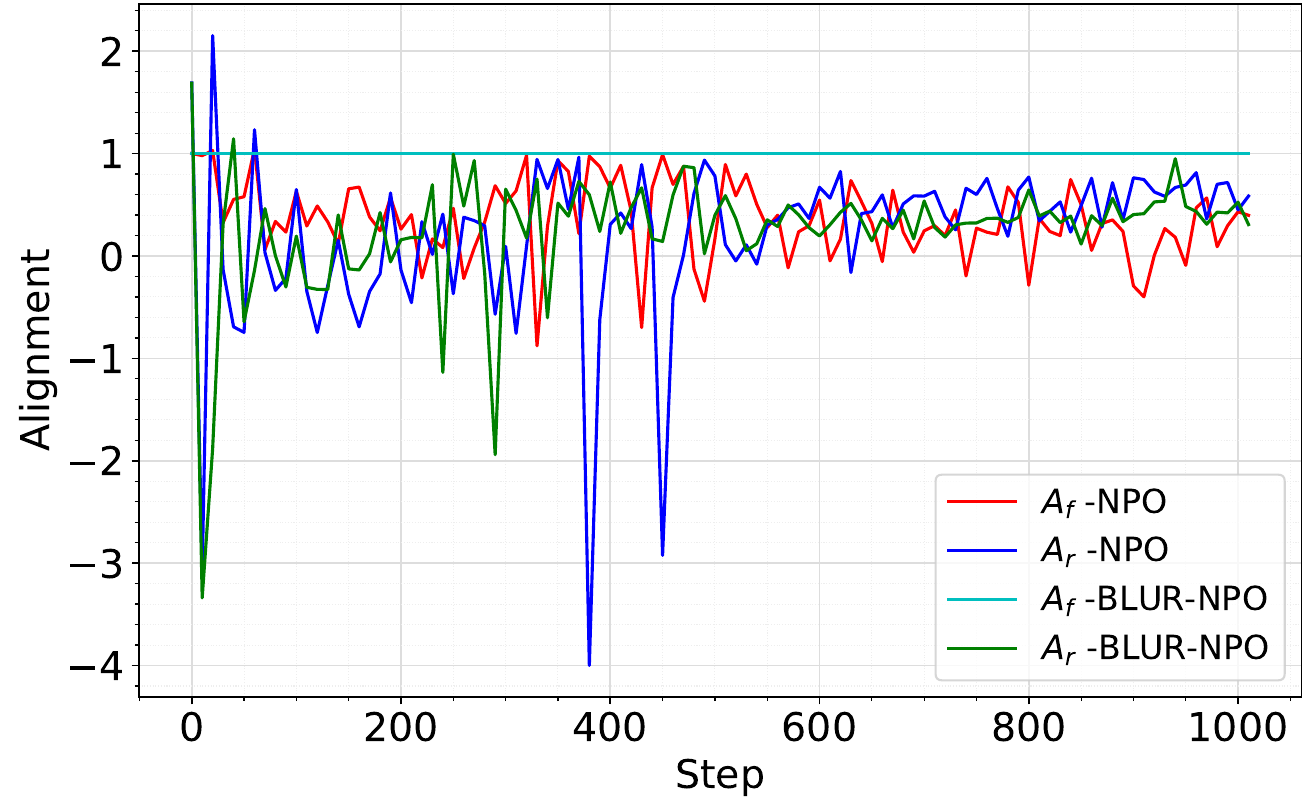}
    \vspace{-0.5cm}
    \caption{Alignment values of forget and retain losses in~\eqref{eq:alig} on MUSE-News using LLaMa2-7B model vs. training step.}
    \label{fig:alig-npo} 
    \vspace{-0.5cm}
\end{wrapfigure}

If $A_f(\bft)$ (resp. $A_r(\bft)$) is positive, it means that the update direction $\hat{u}(\bft)$ will improve the forget loss $f(\bft)$ (resp. retain loss $r(\bft)$). \textbf{Fig~\ref{fig:alig-npo}} plots the change of $A_f(\bft)$ and $A_r(\bft)$ across the training iterations of NPO with  $\lambda=1$. As \textbf{Fig~\ref{fig:alig-npo}} shows, the descent direction $\hat{u}(\bft)$ switches \textit{constantly} the priorities between the forget and retain objectives (note that, in contrast, the proposed algorithm 
\blr always prioritizes the forget loss; see subsequent discussions for details). This example indicates that the regularized formulation with a static $\lambda$ fails to prioritize the loss functions properly and cannot adapt to the complexities of the data and the dynamics of unlearning. Additional experiments with  $\lambda=0.5,1.5,2$, as given in Appendix~\ref{add:exp},~\textbf{Fig~\ref{fig:alig_extra}}, show similar switching behavior. As argued in the introduction, in many practical use cases, the removal of sensitive information, such as copyright-related data or personal information, is critical \cite{gdpr2016general,binns2018fairness}. Therefore, it is useful to have a new unlearning formulation that consistently and explicitly prioritizes one task (e.g., the \textit{forget} task) while treating the other one (e.g., the \textit{retain} task) as an auxiliary task. Towards this end, we exploit a classical optimization paradigm called {\it bi-level optimization}~\cite{maclaurin2015gradient,franceschi2018bilevel,finn2017model,ji2021bilevel,stadie2020learning,zeng2023demonstrations}, which is used to model problems with {\it hierarchical} structure.  In a typical bi-level optimization problem, the upper-level objective function (retain loss, in this case) is minimized over the solution set of a lower-level objective function (forget loss). Therefore, placing more emphasis on the lower-level problem.  More precisely, consider the following problem formulation 
\begin{align}
&\min_{\bft\in \Theta} \; r(\bft) \!\quad\! \text{s.t.} \!\quad\! \Theta \!=\! \arg\min_{\bft\in \R^d} f(\bft).\label{eq:bi-f-l}
\end{align}
In the above formulation, $\Theta$ is the optimal solution set of the lower-level problem. 
We note that~\eqref{eq:bi-f-l} is a specific form of bi-level optimization, often referred to as a {\it simple} bi-level, because the lower-level optimization variable is exactly the same as that of the upper-level ~\cite{sabach2017first,dutta2020algorithms,dempe2021simple}.
It is important to note that  \eqref{eq:bi-f-l} can be viewed as a {\it meta} formulation, where different forget and retain losses can be used to replace the abstract losses $r(\cdot)$ and $f(\cdot)$, e.g., those mentioned in Table \ref{tab:unlearning_methods}. Indeed, in our numerical experiments to be shown shortly, we have demonstrated that it is beneficial to use customized loss functions for certain tasks. 
To illustrate the difference between the bi-level formulation \eqref{eq:bi-f-l} and the weighted sum formulation \eqref{eq:reg_f}, let us consider the following toy example. 
\begin{example}
    Consider a specialization of \eqref{eq:bi-f-l}
\begin{align*}
&\min_{\bft\in \Theta} \left[h(\theta)\!:=\!(\theta\!-\!2)^2 \right], \quad s.t. \quad \Theta \!=\! \arg\min_{\bft\in \R} \left[w(\theta):=|\theta\!-\!1|\!+\!|\theta\!+\!1|\right].  
\end{align*}
Clearly, $w(\theta)$ is minimized over the set $\Theta=[-1,1]$. Thus, the problem is simplified to $ \min_{\bft\in [-1,1]} (\theta-2)^2$ whose optimal solution $\theta^\star=1$. Meanwhile, from \eqref{eq:reg_f},  we can write a regularized optimization problem: 
\begin{align}\label{eq:reg-exmp}
    \min_{\theta\in \R} \left[h(\theta) \!+\! \lambda w(\theta)=(\theta\!-\!2)^2 \!+\! \lambda(|\theta\!-\!1|\!+\!|\theta\!+\!1|) \right]\!.
\end{align}
Note that for an arbitrary choice of $\lambda$, problems~\eqref{eq:reg-exmp}  and \eqref{eq:bi-f-l} are not equivalent. When $\lambda=0$, the problem reduces to $ \min_{\bft\in [-1,1]} (\theta-2)^2$ leading to the optimal solution $\theta=2$, that is outside of $\Theta$. As $\lambda\rightarrow \infty$, the regularization term dominates, forcing the optimal solution to be \textbf{any} value within $\Theta$.
\end{example}
\begin{remark}
    Of course, if necessary,  one can easily switch the order of the lower and upper-level problems, emphasizing more on preserving model utilities. However, we found that both conceptually and numerically, this is not a good modeling choice (at least from the datasets we have tested). Therefore, throughout this paper, we will not mention this case. 
\end{remark} 
In the LLM unlearning, we often deal with nonconvex objectives. When the primal feasibility condition for~\eqref{eq:bi-f-l} cannot be satisfied exactly, we instead aim to converge to a stationary point where $ \|\nabla f(\boldsymbol{\theta})\|^2 \!\leq\! \epsilon_0$ for some $\epsilon_0\geq 0$. Further, we require an approximate stationarity condition of the Lagrangian function $\|\nabla r(\bft) \!+\! \zeta \nabla f(\bft)\|^2 \!\leq\! \epsilon_1$ for some $\epsilon_1\!\geq\! 0$ where $\zeta$ is the Lagrange multiplier. Hence, we aim to find solution $\bft$ that satisfies:
\begin{align}
    & \|\nabla f(\bft)\|^2 \leq \epsilon_0, \label{eq:prim-feas_0} \\
    &  \|\nabla r(\bft) + \zeta \nabla f(\bft)\|^2\leq \epsilon_1. \label{eq:stat}
\end{align}

\section{\blr: Method and Analysis}
\begin{figure}[H] 
\centering
  \includegraphics[width=0.5\linewidth]{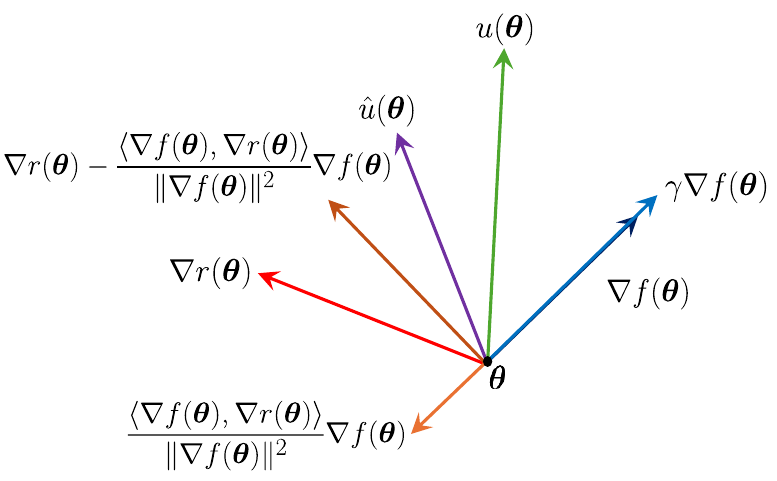}
  \caption{Visualization of the update direction in~\eqref{eq:u_hat} and~\eqref{eq:theta_hat} with their components.}
  \label{fig:2d-update}
  \vspace{-0.5cm}
\end{figure}

In this section, we first discuss the limitations of previously proposed algorithms for solving~\eqref{eq:stat}-\eqref{eq:prim-feas_0}, then we present our scheme, termed Bi-Level UnleaRning (\blr).  Finally, we provide the theoretical guarantees for the proposed algorithm. The majority of the existing works~\cite{sabach2017first,shen2023online,jiang2023conditional,wang2024near} assume that either the upper-level or lower-level objectives is (strongly) convex, but this is certainly not true in the  LLM unlearning setting where both problems are  \textit{non-convex}. Specifically, the scheme proposed in~\cite{sabach2017first} assumes both the lower and upper-level objectives are strongly convex. Moreover, many of them require solving the lower-level problem to some accuracy before updating the upper-level problem, which can incur a significant computational burden when the LLM is large~\cite{franceschi2018bilevel,dagreou2022framework}.
Importantly, we demonstrate that the retain and forget gradients conflict over the unlearning steps, necessitating the design of the update direction in favor of the lower-level objective function.  To this end, we examine the relation between the forget and retain loss gradients across the iterations of the algorithm in~\eqref{eq:theta_hat}. Here, we consider the NPO method, a specific version of the algorithm in~\eqref{eq:theta_hat}, where the forget and retain losses are defined in \textbf{Table~\ref{tab:unlearning_methods}}.  
 \begin{wrapfigure}{r}{0.5\textwidth}
    \centering
    \vspace{-0.3cm}
    \includegraphics[width=\linewidth]{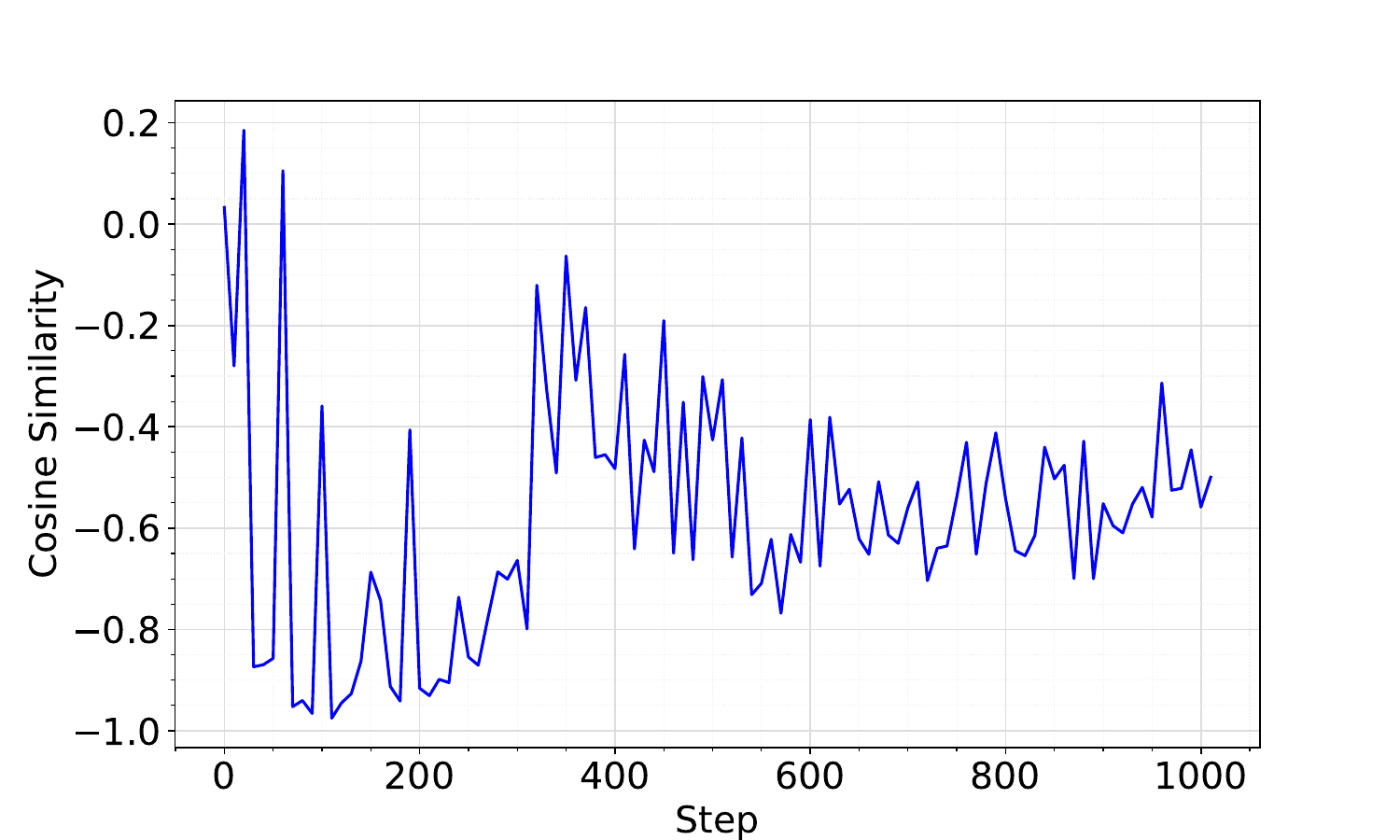}
    \vspace{-0.5cm}
    \caption{Cosine similarity of the gradient forget and retain losses using NPO on MUSE-News dataset and LLaMA2-7B, with $\lambda \!=\!1$ and $\eta\!=\! 10^{-5}$.}
    \label{fig:cos_f_r}
    \vspace{-0.4cm}
\end{wrapfigure}
We conduct an experiment using the NPO method on the MUSE-News dataset with a regularization term of $\lambda=1$ and a learning rate of $\eta=10^{-5}$. 
In Fig. \ref{fig:cos_f_r}, we plot the trajectory of cosine similarity between these two quantities, that is $\frac{\la \nabla f(\bft),\nabla r(\bft)\ra}{\|\nabla f(\bft)\| \|\nabla r(\bft)\|}$.
It is clear that such a similarity measure remains mostly negative, implying that the retain gradient contains a \textit{destructive} component with respect to the forget loss.
More experiments with~$\lambda=0.5, 1.5, 2$, as given in Appendix~\ref{add:exp} and~\textbf{Fig.~\ref{fig:cos_extra}}, further support this observation.
We also run this experiment using the GradDiff method, defined as in \textbf{Table~\ref{tab:unlearning_methods}}, with~$\lambda=0.5, 1, 1.5$ where we observe the similar conflicting pattern in~\textbf{Fig.~\ref{fig:cos_extra_ga}}. 
Thus, naively summing the forget and retain gradients could not provide the desired direction toward minimizing the forget loss (lower-level problem). To ensure convergence to a stationary point of $f$, i.e., $\nabla f(\bft) = 0$, the update direction $u(\bft)$ should move in favor of the objective function $f$. More precisely, the desired update direction $u(\bft)$ should satisfy 
\begin{align*}
    \la \nabla f(\bft), u(\bft) \ra = \gamma \|\nabla f(\bft) \|^2,
\end{align*} 
for some $\gamma > 0$.  To fulfill this condition, we have to appropriately remove {\it destructive} components from the retain gradient. Further, whenever possible, $u(\bft)$ should also contain the non-destructive component of the retain gradient to be able to minimize the upper-level problem.
We propose a novel update direction that satisfies these requirements, given below: 
\begin{align}\label{eq:u}
     u(\bft) = \gamma\nabla f(\bft) \!+\! \nabla r(\bft)\!-\!\frac{\la \nabla f(\bft),\nabla r(\bft)\ra}{\|\nabla f(\bft)\|^2}\nabla f(\bft). 
\end{align}
The update direction in~\eqref{eq:u} can be interpreted in relation to the Gram-Schmidt orthogonalization process. More precisely, the third term on the RHS of~\eqref{eq:u} represents the projection of $\nabla r(\bft)$ onto $\nabla f(\bft)$. The visualization of the update directions in~\eqref{eq:u} and~\eqref{eq:u_hat} with their components is shown in \textbf{Fig.~\ref{fig:2d-update}}.  
With the above discussion about the update direction~\eqref{eq:u}, the proposed algorithm \blr can be simply expressed as as:
\begin{align}
     \bft(t+1)  \!=\! \bft(t) \!-\! \eta \cdot u(\bft(t)), 
     \label{eq:forbun}
\end{align}
where $\bft(0)=\bft_0$ is the pre-trained model; $\eta>0$ is the learning rate. We note that~\blr is a meta-algorithm that can take different forms depending on the specific choices of $f$ and $r$. Subsequently, we use BLUR--$[\cdot]$ to indicate the specific choices of loss functions. We denote BLUR--NPO by using~\blr with the retain loss in~\eqref{eq:cross} and the forget objective in~\eqref{eq:npo}. We plot $A_f(\bft)$ and $A_r(\bft)$ across the optimization steps of BLUR--NPO in~\textbf{Fig.~\ref{fig:alig-npo}}. As we observe,~\blr consistently prioritizes the forget loss over the retain loss at each step.

\subsection{Convergence Analysis}
Next, we present the theoretical analysis of~\blr, demonstrating the convergence behavior of the forget gradient loss $\nabla f(\bft(t))$ and the descent direction $u(\bft(t))$. Towards this end, we make the following assumptions.
\begin{assumption}[\textbf{Function Assumptions}]\label{asm:f}
We assume the following properties on  continuously differentiable functions $f$ and $r$:
\vspace{-0.2cm}
\begin{enumerate}[(a)]
      \item The gradient of $f$ is $\li_f$-Lipschitz, i.e., for any $\bfx,\bfy\in\mathbb{R}^d$ we have $\|\nabla f(\bfx)-\nabla f(\bfy)\|\leq \li_f\|\bfx-\bfy\|$.
      \vspace{-0.2cm}
      \item The gradient of $r$ is $\li_r$-Lipschitz.
      \vspace{-0.1cm}
      \item There exists a constant $C<\infty$ such that $\|f(\bfx)\|, \|r(\bfx)\|, \|\nabla f(\bfx)\|, \|\nabla r(\bfx)\| \leq C$.
 \end{enumerate}
\end{assumption}
We have the following result:
\begin{theorem}\label{thm}
    Under Assumption~\ref{asm:f}, the model generated by using  dynamics in~\eqref{eq:u} -- \eqref{eq:forbun} satisfies
    \begin{align}\label{eq:thm0}
        \frac{1}{T}\sum_{t=0}^{T-1} \|\nabla f(\bft(t)) \|^2 \leq \frac{2C}{T\eta \gamma} + \frac{\li_f}{2\gamma}\eta C^2_1.
    \end{align}
Further, the following holds:
\begin{align}\label{eq:thm1}
        \!\!\frac{1}{T}\sum_{t=0}^{T-1}\!\left[\|\nabla f(\bft(t)) \|^2\!+\!\|u(\bft(t))\|^2\right] \!\leq \!\frac{4C}{\eta T}\left(\frac{1}{2\gamma}\!+\!1\!+\!\gamma\right) \!+\! \li_f\eta C^2_1 \left(\frac{1}{2\gamma}\!+\!\gamma\right) \!+\! 2\gamma C_1 \sqrt{\frac{\li_f}{\gamma}\eta},
    \end{align}
     for every $T\geq \frac{4C}{\li_f C^2_1 \eta^2}$ where $C_1:=(2+\gamma)C$.
\end{theorem}
The proof of Theorem~\ref{thm} is presented in Appendix~\ref{ap:proof_thm}. Intuitively, $\|u(\cdot)\|^2$ quantifies the degree of conflict between $\nabla f$ and $\nabla r$; it represents how much we can reduce $f$ without increasing $r$. 
\begin{remark}
To maximize the upper bounds~\eqref{eq:thm0} and~\eqref{eq:thm1}, 
it can be verified that we should choose $\eta=\frac{2}{C_1}\sqrt{\frac{C}{\li_f}} \frac{1}{T^{\frac{1}{2}}}$ for any $\gamma>\max\left(0, \frac{2}{\sqrt{\li_f C}} - 2\right)$. Plugging these into~\eqref{eq:thm0} and~\eqref{eq:thm1}, we conclude 
\begin{align}
    & \!\frac{1}{T}\sum_{t=0}^{T-1} \|\nabla f(\bft(t)) \|^2 \leq \frac{C_1\sqrt{\li_f C}}{\gamma} \frac{1}{T^{\frac{1}{2}}}, \label{eq:norm_nabla_f}\\
    &\! \frac{1}{T}\sum_{t=0}^{T-1}\!\left[\|\nabla f(\bft(t)) \|^2\!+\!\|u(\bft(t))\|^2\right]\!\leq \! 2 C_1\sqrt{\li_f C}\left(\frac{1}{\gamma}\!+\!1\!+\!2\gamma\right) \!\frac{1}{T^{\frac{1}{2}}} \!+\! 2\sqrt{2C_1\gamma} (\li_f C)^{\frac{1}{4}} \frac{1}{T^{\frac{1}{4}}}\!,\label{eq:norm_u}.
\end{align}
Our results show that with a proper choice of step size,  the temporal average of the norm of the forget gradient and the update direction decreases at a rate of $\mathcal{O}(T^{-1/2})$ and $\mathcal{O}(T^{-1/4})$, respectively.
\end{remark}
\begin{remark}
Using~\eqref{eq:norm_nabla_f}, we get $\lim_{t\rightarrow \infty} \nabla f(\bft(t))=0$ which indicates convergence to a stationary point of the lower-level problem, i.e.,~\eqref{eq:prim-feas_0} is satisfied. Further, from~\eqref{eq:norm_u}, we have $\lim_{t\rightarrow \infty} u(\bft(t)) =  0$, i.e., $\lim_{t\rightarrow \infty} \left[ \nabla r(\bft(t)) + \left(\gamma - \frac{\la \nabla f(\bft(t)),\nabla r(\bft(t))\ra}{\|\nabla f(\bft(t))\|^2} \right)\nabla f(\bft(t))\right]=0$. This implies that the stationary condition in~\eqref{eq:stat} is satisfied with $\zeta=\gamma - \frac{\la \nabla f(\bft(t)),\nabla r(\bft(t))\ra}{\|\nabla f(\bft(t))\|^2}$; thus, both desired optimality conditions are fulfilled. Moreover, in practice, we observe that the gradients of the forget and retain losses are often conflicting (see \textbf{Fig.~\ref{fig:cos_f_r}}), i.e., $\la \nabla f(\bft(t)),\nabla r(\bft(t))\ra \leq 0$. Consequently, the term $\zeta$ remains \textit{non-negative}, ensuring that the resulting stationary point satisfies the complementary condition in~\eqref{eq:stat}.
\end{remark}

\vspace{-0.45cm}
\section{Experiment}\label{sec:exp}
\vspace{-0.25cm}
In this section, we evaluate the performance of the proposed algorithm,~\blr, and other state-of-the-art unlearning methods.

\noindent{\textbf{Unlearning Tasks.}}
We test unlearning algorithms on three popular benchmark datasets TOFU~\cite{maini2024tofu}, MUSE~\cite{shi2024muse}, and WMDP~\cite{li2024wmdp}. The TOFU dataset contains 200 fictitious author profiles, each including 20 question-answer pairs generated by GPT-4 using predefined attributes. Here, we consider forget05 and forget10 scenarios, representing $5\%$ and $10\%$ forget sets, respectively. The MUSE dataset consists of two corpora, namely, Harry Potter books (Books) and news articles (News). The WMDP benchmark is developed for knowledge-based unlearning to remove hazardous knowledge in biosecurity, cybersecurity,
and chemical security. We conduct the experiments on the TOFU dataset using the public fine-tuned LLaMA-3.2-1B-Instruct model\footnote{\url{https://huggingface.co/open-unlearning/tofu_Llama-3.2-1B-Instruct_full}}. Further, we run simulations on the MUSE benchmark using the public fine-tuned LLaMA2-7B model\footnote{\url{https://huggingface.co/muse-bench/MUSE-news_target}\par \url{https://huggingface.co/muse-bench/MUSE-books_target}}. Finally, we exploit the Zephyr-7B-beta model\footnote{\url{https://huggingface.co/HuggingFaceH4/zephyr-7b-beta}} for WMDP.

\noindent{\textbf{LLM Unlearning Methods.}} 
We use `\texttt{Original}' to indicate the fine-tuned model using the TOFU/MUSE datasets and the pre-trained model for WMDP. We use `\texttt{Retrain}' to indicate models retrained while {\it excluding} the forget set; such a `\texttt{Retrain}' model is considered the gold standard for unlearning, and such a model is available for the TOFU and MUSE benchmarks. 

For the TOFU dataset, we compare the performance of NLUR--NPO against GA, GradDiff, NPO, and SimNPO. We evaluate BLUR-NPO and compare its performance against GA, GradDiff, NPO, SimNPO, and NGDiff on the MUSE dataset. For the WMDP dataset, we compare BLUR--NPO with the Representation misdirection for unlearning (RMU) developed in~\cite{li2024wmdp} that directs the representations of forget samples toward random representations while preserving the representations of retain samples. The forget and retain losses are defined as follows
\begin{subequations}\label{eq:rmu}
    \begin{align}
   &\ell_{\text{RMU},r}(y|x;\bft) =   \|M_i(x; \bft) - M_i(x; \bft_0)\|_2^2,\label{eq:rmu_r}\\
   &\ell_{\text{RMU},f}(y|x;\bft) =  \|M_i(x; \bft) - c \cdot \mathbf{u}\|_2^2,\label{eq:rmu_f}
\end{align}
\end{subequations}
where $M_i(x; \bft)$ is a function that returns the hidden representation of $\bft$ at some layer $i$, and a fixed random unit vector $\bfu$ sampled uniformly from $[0,1)$. Here, $c$ is a hyperparameter that controls activation scaling.
Utilizing the gradient direction in~\eqref{eq:u_hat} with the retain loss~\eqref{eq:rmu_r} and the forget objective~\eqref{eq:rmu_f} to solve the regularized optimization problem~\eqref{eq:reg_f} is referred to as the RMU unlearning method. We evaluate~\blr with the same retain loss and forget objective as expressed in \eqref{eq:rmu}, resulting in a new unlearning method referred to as BLUR--RMU. Experiments are conducted on the WMDP benchmark, comparing RMU, NPO, SimNPO, BLUR--NPO, and BLUR--RMU. Additionally, we implement NGDiff method with the retain loss and forget objectives in \eqref{eq:rmu}, termed NGDiff--RMU.

\begin{table}[t]
\centering
\caption{Performance of various unlearning methods on the TOFU benchmark using the LLaMA-3.2-1B-Instruct model for forget05 and forget10.}
\resizebox{0.75\textwidth}{!}{ 
\begin{tabular}{c|cc|cc|cc}
\toprule
\multirow{3}{*}{\textbf{Method}} 
& \multicolumn{2}{c|}{\textbf{Forget Quality} $\uparrow$} 
& \multicolumn{2}{c|}{\textbf{Forget Truth Ratio} $\uparrow$} 
& \multicolumn{2}{c}{\textbf{Model Utility} $\uparrow$} \\
\cmidrule(lr){2-3} \cmidrule(lr){4-5} \cmidrule(lr){6-7}
& forget05 & forget10 & forget05 & forget10 & forget05 & forget10 \\
\midrule
Original   & 2.96e--13 & 8.08e--22 & 0.47 & 0.48 & 0.60 & 0.60 \\
Retain     & 1.00      & 1.00      & 0.63 & 0.63 & 0.60 & 0.59 \\
\midrule
GA         & 1.94e--119 & 1.06e--239 & 2.97e--24 & 9.82e--26 & 0.00 & 0.00 \\
GradDiff   & 1.94e--119 & 3.76e--219 & 8.13e--9 & 0.002 & 0.52 & 0.49 \\
NPO        & 0.40       & 0.08       & \textbf{0.70} & \textbf{0.65} & 0.47 & 0.51 \\
SimNPO     & 0.068      & 0.005      & 0.52 & 0.52 & \textbf{0.57} & \textbf{0.58} \\
\rowcolor{blue!20}
BLUR--NPO  & \textbf{0.80} & \textbf{0.91} & 0.68 & 0.64 & 0.52 & 0.55 \\
\bottomrule
\end{tabular}
}
\vspace{-5mm}
\label{tab:tofu}
\end{table}

\noindent{\textbf{Evaluation Metrics.}}
We list the metrics to evaluate the performance of each unlearning task.

\underline{TOFU.} We measure the \textit{forget quality}, which assesses how well the unlearned model mimics the \texttt{Retain} model. \textit{Model utility} captures the general capabilities and real-world knowledge retained by the model after unlearning. Also, we report the \textit{truth ratio}, which shows how likely the model is to select the correct answer over an incorrect one.

\underline{MUSE.} After the unlearning phase, we expect to satisfy four key criteria: (1) \textit{No verbatim memorization}: The model should no longer be able to generate exact substrings or sequences that match any content from the forget set. (2) \textit{No knowledge memorization on the forget set}: The model should not be able to generate accurate or meaningful answers to questions regarding the forget set. (3) \textit{No privacy leakage}: There should be no indication that the model was ever trained on the forget set. (4) \textit{High knowledge memorization on the retain set}: The model should maintain good performance on the retain set, ensuring a high value for knowledge memorization on the retain set.

\underline{WMDP.} To evaluate the model's performance on the forget and retain datasets, we use the Question-Answering (QA) technique, which involves assessing the accuracy of the model's answers based on a given corpus. \texttt{Bio. Acc.} and \texttt{Cyber Acc.} represent the model's accuracy on WMDP-Bio and WMDP-Cyber, respectively; \texttt{MMLU} contains subjects that should not be unlearned where performance is also evaluated using the QA technique. 

Overall, we can categorize these metrics into two classes: (1) \textit{Unlearning Effectiveness:} This class evaluates how effectively undesired data and its influences are removed from model capabilities. (2) \textit{Utility Preservation:} Metrics in this class assess the model's performance on standard utility tasks after the unlearning phase. {{More
implementation details are provided in Appendix~\ref{app:hpyer}.}}

\noindent{\textbf{Results.}} Our results are summarized below.

\underline{TOFU.} \textbf{Table~\ref{tab:tofu}} compares the performance of BLUR--NPO with other unlearning baselines on TOFU forget05 and forget10. As shown, BLUR--NPO achieves \textit{outstanding} performance in forget quality ($0.4$ to $0.8$ for forget05 and $0.08$ to $0.91$ for forget10) while maintaining comparable results in forget truth ratio and model utility.

\underline{MUSE.}
We compare the performance of BLUR--NPO with other baselines in \textbf{Table~\ref{tab:muse}} on both the MUSE News and Books datasets. Clearly, BLUR--NPO outperforms all baselines across \textit{all} unlearning efficiency metrics, i.e., verbatim memorization (VerbMem) on the forget set, knowledge memorization (KnowMem) on the forget set, and KnowMem on the retain set. Moreover, it achieves no VerbMem on the forget set as GA, NGDiff, and NPO methods. In~\textbf{Fig.~\ref{fig:ab}}, we present an ablation study on hyperparameters for the News corpus. To highlight the BLUR--NPO's performance on the Books corpus,~\textbf{Fig.~\ref{fig:trade-off_books}} shows the trade-off between KnowMem on the forget and retain datasets.

To understand how the performance of different algorithms evolves over the iterations, \textbf{Fig.~\ref{fig:verb_knw}} plots the trajectory of the VerbMem on the forget set (lower values are better) and KnowMem on the retain set (higher values are better), for the MUSE-News unlearning task. As the plots show, GA, NGDiff, NPO, and BLUR--NPO exhibit no VerbMem on the forget set, while the other baselines fail to achieve the complete no-VerbMem status. Further, BLUR--NPO achieves consistently high levels of KnowMem across all epochs. NGDiff, NPO, and SimNPO exhibit better model utility than GradDiff and GA but are still less effective than BLUR--NPO. Moreover, it is interesting to observe that BLUR--NPO first prioritizes optimizing the forget performance by aggressively reducing VerbMem to zero. Then, as the model has additional capacity, it gradually increases the KnowMem on the retain set as optimization goes.

In~\textbf{Table~\ref{tab:muse_news}}, we provide examples of model responses after unlearning with different schemes, compared to the ground truth. For the forget set, the goal is to unlearn the \textit{Answer} while for the retain set, the objective is to generate a response that matches the \textit{Answer}. These visualizations show that BLUR--NPO delivers the best results, accurately responding to questions from the retain set while providing non-informative responses to questions from the forget set. However, all other unlearning methods either fail to provide fully correct answers for the retain set or are unable to decline answering questions from the forget set. 
\begin{table}
\centering
\caption{Performance of various unlearning methods on the MUSE benchmark using the News and Books corpora with the LLaMA2-7B model.}
\vspace{1mm}
\resizebox{0.8\textwidth}{!}{ 
\begin{tabular}{c|cc|cc|cc|cc}
\toprule[1pt]
\multirow{3}{*}{\textbf{Method}} 
& \multicolumn{2}{c|}{\textbf{VerbMem on $\mathcal{D}_f$ $\downarrow$}} 
& \multicolumn{2}{c|}{\textbf{KnowMem on $\mathcal{D}_f$ $\downarrow$}} 
& \multicolumn{2}{c|}{\textbf{PrivLeak $\rightarrow 0$}} 
& \multicolumn{2}{c}{\textbf{KnowMem on $\mathcal{D}_r$ $\uparrow$}} \\
\cmidrule{2-9}
& News & Books & News & Books & News & Books & News & Books \\
\midrule
Original     & 58.4 & 99.8 & 63.9 & 59.4 & -99.8 & -57.5 & 55.2 & 66.9 \\
Retrain      & 20.8 & 14.3 & 33.1 & 28.9 & 0.0 & 0.0 & 55.0 & 74.5 \\
\midrule
GA           & 0.0  & 0.0  & 0.0  & 0.0  & 5.2  & -23.6 & 0.0  & 0.0  \\
GradDiff     & 4.9  & 0.0  & 31.3 & 0.0  & 107.9 & -24.1 & 22.9 & 14.4 \\
NGDiff       & 0.0  & 0.0  & 35.7 & 0.0  & 109.5 & -20.6 & 41.8 & 44.1 \\
NPO          & 0.0  & 0.0  & 43.9 & 0.0  & 109.4 & -30.3 & 37.5 & 31.8 \\
SimNPO       & 6.7  & 0.0  & 46.2 & 0.0  & 62.6 & -24.2 & 35.9 & 49.3 \\
\midrule
\rowcolor{blue!20}
BLUR--NPO    & \textbf{0.0}  & 0.0  & \textbf{29.0} & 0.0  & 109.5 & -22.6 & \textbf{46.7} & \textbf{52.7} \\
\bottomrule[1pt]
\end{tabular}
}
\label{tab:muse}
\vspace*{-2mm}
\end{table}

\underline{WMDP.} 
\textbf{Table~\ref{tab:wmdp}} compares our algorithms, BLUR--NPO and BLUR--RMU, with RMU, NPO, SimNPO, and NGDiff--RMU on the WMDP benchmark. Recall that BLUR--NPO is the specification of the retain loss in~\eqref{eq:cross} and the forget objective in~\eqref{eq:npo}. Also, the BLUR--RMU and NGDiff--RMU leverage the loss function developed in the RMU algorithm.  As shown, BLUR--NPO and BLUR--RMU outperform \textit{all} other methods in unlearning efficiency. BLUR--NPO performs better than NPO and SimNPO but is less effective than RMU in utility preservation. Interestingly, BLUR--RMU achieves roughly the same performance as RMU and NGDiff--RMU on the retain set, while significantly outperforming RMU on unlearning efficacy. These results highlight the effectiveness of our approach relative to previously proposed algorithms. To visualize better, we compute the average accuracy of WMDP-Bio and WMDP-Cyber and then plot it vs. the accuracy of MMLU in~\textbf{Fig.~\ref{fig:trade-off_wmdp}}.
\vspace{-0.25cm}
\section{Conclusion}
\vspace{-0.25cm}
\begin{wraptable}{r}{0.5\textwidth}
\vspace{-0.5cm}
\caption{Performance of various unlearning methods on the WMDP benchmark.}
\vspace{-0.1cm}
\renewcommand{\arraystretch}{1.0}
\resizebox{0.5\textwidth}{!}{
\begin{tabular}{c|cc|c}
\toprule[1pt]
\multirow{2}{*}{\textbf{Method}} & \multicolumn{2}{c|}{\textbf{Unlearning Efficacy}} & \textbf{Utility Preservation} \\
\cmidrule{2-4}
& Bio. Acc. \textbf{$\downarrow$} & Cyber Acc. \textbf{$\downarrow$} & MMLU \textbf{$\uparrow$} \\
\midrule
Original & 63.7 & 44.0 & 58.1 \\
\midrule
RMU & 31.2 & 28.2 & 57.1 \\
NPO & 42.5 & 28.3 & 40.0 \\
SimNPO & 41.6 & 32.2 & 47.1 \\
NGDiff--RMU & 35.6 & 26.8 & 57.4 \\
\midrule
\rowcolor{blue!20}
BLUR--NPO & 27.6 & \textbf{26.5} & 48.4 \\
\rowcolor{blue!20}
BLUR--RMU & \textbf{26.9} & 26.6 & 57.0 \\
\bottomrule
\end{tabular}
}
\label{tab:wmdp}
\end{wraptable}
In this paper, we introduced a new LLM unlearning framework based on a bi-level optimization approach that prioritizes the forget loss over the retain objective in a hierarchical structure. To solve the proposed optimization problem, we developed a novel algorithm, termed~\blr. Then, we provided the theoretical analysis for the non-convex setting. Our extensive experiments on various LLM unlearning tasks demonstrated that our approach outperformed all the state-of-the-art algorithms. For future work, we will explore higher-order algorithms to further enhance unlearning efficiency and effectiveness.


\bibliography{ref}
\bibliographystyle{plain}

\appendix
\counterwithin{figure}{section}

\section{The Preliminaries}
\begin{lemma}\label{lm:aux1}
For any $\bft \in \R^d$, we can write
  \begin{enumerate}[(a)]
      \item $\la \nabla f(\bft), u(\bft) \ra =  \gamma \|\nabla f(\bft) \|^2$.
  \end{enumerate}
    Also, under Assumption~\ref{asm:f}, for every $\bft \in \R^d$, we get
\begin{enumerate}[(a)]
    \setcounter{enumi}{1}
        \item $\| u(\bft) \| \leq C_1$,
        \item $\xi(\bft) \|\nabla f(\bft)\|^2 \leq \gamma\|\nabla f(\bft) \|^2 + C\| \nabla f(\bft)\|$,
        \item $ \xi(\bft) \la \nabla f(\bft) , u(\bft)\ra \leq \gamma^2\|\nabla f(\bft) \|^2 + C\gamma\| \nabla f(\bft)\|$,
        \item $\la \nabla r(\bft), u(\bft) \ra \geq \| u(\bft) \|^2  - \gamma^2\|\nabla f(\bft) \|^2 - C\gamma\| \nabla f(\bft)\|$,
    \end{enumerate}
\end{lemma}
where $\xi(\bft):=\frac{\gamma\|\nabla f(\bft)\|^2 - \langle \nabla f(\bft), \nabla r(\bft) \rangle}{\|\nabla f(\bft)\|^2}$.
\begin{proof}
From the definition of $\xi(\bft)$ and $u(\bft)$ in~\eqref{eq:u}, we have
        \begin{align*}
             \la \nabla f(\bft), u(\bft) \ra & = \la \nabla f(\bft),\nabla r(\bft)\ra + \xi(\bft) \|\nabla f(\bft) \|^2\nonumber\\
             & = \la \nabla f(\bft),\nabla r(\bft)\ra + \gamma\|\nabla f(\bft)\|^2 - \langle \nabla r(\bft), \nabla r(\bft) \rangle = \gamma \|\nabla f(\bft) \|^2, 
        \end{align*}
        that completes the proof of part (a).
    From the definition of $u(\bft)$ and the triangle inequality, we can write
    \begin{align*}
        \| \xi(\bft) \nabla f(\bft) + \nabla r(\bft) \| &\leq |\xi(\bft)|\| \nabla f(\bft)\| + \|\nabla r(\bft) \|\nonumber\\
        & \leq \gamma\| \nabla f(\bft)\| + \frac{|\la \nabla f(\bft),\nabla r(\bft)\ra|}{\| \nabla f(\bft)\|}+\| \nabla r(\bft)\|\nonumber\\
        & \stackrel{\rm{(a)}}{\leq} \gamma\| \nabla f(\bft)\|+2\| \nabla r(\bft)\| \leq (2+\gamma)C,
    \end{align*}
    where $\rm{(a)}$ follows from the Cauchy-Schwartz inequality and the last step holds due to the Assumption~\ref{asm:f}.

    To prove part (c), using the Cauchy-Schwartz inequality, we get
    \begin{align}\label{eq:lm_b}
        \xi(\bft) \|\nabla f(\bft)\|^2 &= \gamma\|\nabla f(\bft)\|^2 \!-\! \langle \nabla f(\bft), \nabla r(\bft) \rangle\nonumber\\
        & \leq \gamma\|\nabla f(\bft)\|^2 \!+\! \|\nabla f(\bft)\|\cdot \|\nabla r(\bft)\| \nonumber\\
        &\leq \gamma\|\nabla f(\bft)\|^2 \!+\! C \|\nabla f(\bft)\|,
    \end{align}
    where the last inequality holds from Assumption~\ref{asm:f}. Now, we prove part (d). From~\eqref{eq:u}, we have
    \begin{align}\label{eq:lm2_0}
        \xi(\bft) \la \nabla f(\bft) , u(\bft)\ra  &= \xi(\bft) \la \nabla f(\bft), \xi(\bft) \nabla f(\bft)+\nabla r(\bft)\ra\nonumber\\
        & = \xi(\bft) \left(  \xi(\bft) \|\nabla f(\bft) \|^2 + \la \nabla f(\bft),\nabla r(\bft)\ra \right)\nonumber\\
        & = \xi(\bft) \left( \gamma \| \nabla f(\bft)\|^2 - \la \nabla f(\bft),\nabla r(\bft)\ra + \la \nabla f(\bft),\nabla r(\bft)\ra \right)\nonumber\\
        & = \xi(\bft) \gamma \| \nabla f(\bft)\|^2.
    \end{align}
    This together with part (c) leads us to part (d). Finally, to show part (e), from part (d), we get
    \begin{align*}
        \la \nabla r(\bft), u(\bft) \ra &= \la u(\bft) - \xi(\bft) \nabla f(\bft), u(\bft) \ra\nonumber\\
        & \!=\! \| u(\bft) \|^2 \!-\! \xi(\bft) \la \nabla f(\bft) , u(\bft)\ra \!\geq\! \| u(\bft) \|^2  \!-\! \gamma^2\|\nabla f(\bft) \|^2 \!-\! C\gamma\| \nabla f(\bft)\|.
    \end{align*}
\end{proof}
\begin{lemma}\label{lm:aux2}
     For any vectors $\{\bfu_i\}_{i=1}^{n}\in \R^d$, we get $ \lnr \sum_{i=1}^{n} \bfu_i \rnr^2 \leq n \sum_{i=1}^{n} \lnr \bfu_i \rnr^2$.
\end{lemma}
\section{Proof of Theorem~\ref{thm}}\label{ap:proof_thm}
Using the fact that $f$ is $\li_f$-Lipschitz from Assumption~\ref{asm:f}, we get
    \begin{align}\label{eq:g_t_1}
        f(\bft(t\!+\!1)) \!-\! f(\bft(t)) &\leq \la \nabla f(\bft(t)),\bft(t+1)-\bft(t)\ra + \frac{\li_f}{2}\|\bft(t+1)-\bft(t)\|^2\nonumber\\
        & = -\eta \la \nabla f(\bft(t)), u(\bft(t))\ra + \frac{\li_f}{2}\eta^2 \|u(\bft(t)) \|^2\nonumber\\
        & \stackrel{\rm{(a)}}{=} -\eta \gamma \|\nabla f(\bft(t)) \|^2 \!+\! \frac{\li_f}{2}\eta^2 \|u(\bft(t)) \|^2  \leq \!-\!\eta \gamma \|\nabla f(\bft(t)) \|^2 \!+\! \frac{\li_f}{2}\eta^2 C^2_1, 
    \end{align}
where $\rm{(a)}$ follows from Lemma~\ref{lm:aux1}-(a), and the last step holds due to Lemma~\ref{lm:aux1}-(b). 
Applying a telescopic summation in~\eqref{eq:g_t_1}, we can write
\begin{align*}
    \frac{1}{T}\sum_{t=0}^{T-1} \|\nabla f(\bft(t)) \|^2 & \leq \frac{1}{T\eta \gamma}\left(f(\bft(0))-f(\bft(T))\right) + \frac{\li_f}{2\gamma}\eta C^2_1  \leq \frac{2C}{T\eta \gamma} \!+\! \frac{\li_f}{2\gamma}\eta C^2_1,
\end{align*}
where the last step follows from Assumption~\ref{asm:f}. That completes the proof of the first claim of the theorem. Similarly, since $r$ is $\li_r$-Lipschitz from Assumption~\ref{asm:f}, we can write
\begin{align}\label{eq:f_x_2}
    r(\bft(t+1)) - r(\bft(t)) &\leq \la \nabla r(\bft(t)),\bft(t+1)-\bft(t)\ra + \frac{\li_r}{2}\|\bft(t+1)-\bft(t)\|^2\nonumber\\
    & = -\eta\| u(\bft(t))\|^2 + \eta \xi(\bft(t)) \la \nabla f(\bft(t)), u(\bft(t))\ra + \frac{\li_r}{2}\eta^2\|u(\bft(t)) \|^2\nonumber\\
    & \leq -\frac{\eta}{2} \| u(\bft(t)) \|^2 + \eta \left( \gamma^2\|\nabla f(\bft(t)) \|^2 + C\gamma\| \nabla f(\bft(t))\|\right).
\end{align}
Applying a telescopic summation in~\eqref{eq:f_x_2}, we arrive at
\begin{align*}
    \frac{1}{T}\sum_{t=0}^{T-1} \|u(\bft(t))\|^2 & \leq \frac{2}{\eta T}(r(\bft(0))-r(\bft(t))) + \frac{2}{T}\left(\gamma^2\sum_{t=0}^{T-1}\| \nabla f(\bft(t))\|^2 + \gamma\sum_{t=0}^{T-1}\| \nabla f(\bft(t))\|\right)\nonumber\\
    & \leq \frac{4C}{\eta T} + \frac{2}{T}\left(\gamma^2\sum_{t=0}^{T-1}\| \nabla f(\bft(t))\|^2 + \gamma\sum_{t=0}^{T-1}\| \nabla f(\bft(t))\|\right)\nonumber\\
    & \stackrel{\rm{(a)}}{\leq} \frac{4C}{\eta T} + \frac{2}{T}\left(\gamma^2\sum_{t=0}^{T-1}\| \nabla f(\bft(t))\|^2 + \gamma\sqrt{T}\sqrt{\sum_{t=0}^{T-1}\| \nabla f(\bft(t))\|^2}\right)\nonumber\\
    & \stackrel{\rm{(b)}}{\leq}  \frac{4C}{\eta T} + \frac{2}{T}\left(\frac{2C\gamma}{\eta} + \frac{\li_f\gamma}{2}\eta C^2_1 T + \gamma\sqrt{T}\sqrt{\frac{2C}{\eta \gamma} + \frac{\li_f}{2\gamma}\eta C^2_1 T} \right)\nonumber\\
    & \leq \frac{4C}{\eta T}(1+\gamma) + \li_f \gamma \eta C^2_1 + 2\gamma C_1 \sqrt{\frac{\li_f}{\gamma}\eta},
\end{align*}
where $\rm{(a)}$ follows from Lemma~\ref{lm:aux2}, step $\rm{(b)}$ holds due to~\eqref{eq:thm0}, and finally the last inequality can be concluded from $T\geq \frac{4C}{\li_f C^2_1 \eta^2}$.
\section{Additional Experiment Results and Details}\label{add:exp}
\subsection{Experiment Setups}\label{app:hpyer}
\noindent{\textbf{Computational Configurations.}} All experiments are conducted on 8 NVIDIA A100 GPUs. Next, we present the hyperparameters used in each unlearning task, TOFU, MUSE, and WMDP.

\underline{TOFU.} We conduct experiments for 10 epochs using a learning rate of $10^{-5}$ and a batch size of 32. A grid search is performed over the range $[0.5, 2]$ for $\gamma$, and the hyperparameter $\beta$ is searched within $[0.05, 0.2]$, with the final value of $0.1$. We set $\lambda = 1$ for GradDiff and NPO, and $\lambda = 2.5$ for SimNPO. The hyperparameters $\beta$ for NPO and SimNPO are set to $0.1$ and $2.5$, respectively. For the SimNPO unlearning scheme, the hyperparameter $\alpha$ is fixed at $0.125$. We summarized the hyperparameters in \textbf{Table~\ref{tab:hyper_tofu}}.
\begin{wraptable}{r}{0.5\textwidth}
    \resizebox{0.4\textwidth}{!}{
    \begin{tabular}{l|c|c|c|c|c}
        \toprule[1pt]
        \textbf{Method} & \textbf{$\eta$} & \textbf{$\beta$} & \textbf{$\gamma$} & \textbf{$\alpha$} & \textbf{$\lambda$} \\
        \midrule
        GA  & $10^{-5}$ & - & - & - & - \\
        \midrule
        GradDiff  & $10^{-5}$ & - & - & - & 1.0 \\
        \midrule
        NPO  & $10^{-5}$ & 0.1  & -  & - & 1.0 \\
        \midrule
        SimNPO  & $10^{-5}$ & 2.5  & -  & 2.0 & 0.15 \\
        \midrule
        BLUR--NPO  & $10^{-5}$ & 0.1  & 1.0  & - & - \\
        \bottomrule
    \end{tabular}
    }
    \caption{Hyperparamters for various unlearning methods on TOFU benchmark}
    \label{tab:hyper_tofu}
\end{wraptable}
\underline{MUSE.} We train our algorithm BLUR--NPO for $10$ epochs with a constant learning rate of $2.5\times 10^{-5}$ for the news dataset and $10^{-5}$ for the books dataset, the batch size of $32$, and an input length of $2,048$ tokens. We perform a grid search for $\gamma$ in the range of $[0.8, 1.2]$ and set $\gamma = 1.0$ as the final value. For the NPO loss in~\eqref{eq:npo}, $\beta$ is set to $0.05$ for the news dataset and $0.4$ for the books dataset. The evaluation pipelines strictly follow the setup detailed by~\cite{shi2024muse}. Further, we evaluate the model's performance after the unlearning process and select the optimal model as the final result. We use a constant learning rate of $2.5 \times 10^{-5}$ for news and $5 \times 10^{-6}$ for books datasets in NGDiff, with $\beta=0.1$ for both corpora. The hyperparameters for other unlearning methods GA, GradDiff, NPO, and SimNPO are set according to the works of~\cite{shi2024muse,fan2024simplicity}. \textbf{Table~\ref{tab:hyper_muse}} summarizes our method's optimal combination of the hyperparameters, determined through grid search. We used the hyperparameters reported in the corresponding papers for other unlearning methods.
\begin{table}[h]
    \centering
    \resizebox{0.52\textwidth}{!}{ 
    \begin{tabular}{l|c|c|c|c|c|c}
        \toprule[1pt]
        \textbf{Method} & \textbf{Dataset} & \textbf{$\eta$} & \textbf{$\beta$} & \textbf{$\gamma$} & \textbf{$\alpha$} & \textbf{$\lambda$} \\
        \midrule
        GA & News/Books  & $10^{-5}$ & -  & -  & - & 0 \\
        \midrule
        GradDiff & News/Books  & $10^{-5}$ & -  & -  & - & 1 \\
        \midrule
        \multirow{2}{*}{NGDiff} & News  & $2.5\times 10^{-5}$ & 0.1  & -  & - & - \\
                                & Books & $5\times 10^{-6}$   & 0.1  & -  & - & - \\
        \midrule
        NPO & News/Books  & $10^{-5}$ & -  & -  & - & 1 \\
        
        \midrule
        \multirow{2}{*}{SimNPO} & News  & $10^{-5}$ & 0.7  & -  & 0.0 & 0.1  \\
                                & Books & $10^{-5}$ & 0.75  & -  & 0.0 & 0.1  \\
        \midrule
        \multirow{2}{*}{BLUR--NPO} & News  & $2.5\times 10^{-5}$ & 0.05 & 1.0 & - & - \\
                                   & Books & $10^{-5}$           & 0.4 & 1.0 & - & -  \\
        \bottomrule
    \end{tabular}
    }
    \caption{Hyperparameters for various unlearning methods on MUSE benchmark}
    \label{tab:hyper_muse}
\end{table}

\noindent\underline{WMDP.} We run the experiments for BLUR--NPO using a constant learning rate of $2\times 10^{-6}$ and a batch size of $4$. A grid search is performed for $\gamma$ within the range $[0.5, 1.5]$, and the parameter $\beta$ in~\eqref{eq:npo} is explored within the range $[0.001, 0.01]$. As for the RMU and BLUR--RMU algorithms, we follow the implementation details provided in WMDP~\cite{li2024wmdp}. More precisely, for BLUR--RMU, we conduct a grid search for the parameter $\gamma$ over the range $[0.001,0.002]$ and use 0.00125 as the final $\gamma$. We also employ a constant learning rate of $4\times 10^{-2}$ and train for 150 steps for BLUR--RMU. We exploit \texttt{lm-evaluation-harness v0.4.2}~\cite{gao2021framework} to standardize prompts. We set the experimental parameters for other unlearning methods, RMU, NPO, and SimNPO, as described in~\cite{li2024wmdp,jia2024wagle,fan2024simplicity}. We set a fine-tuned and constant learning rate of $10^{-5}$ for NGDiff--RMU method. \textbf{Table~\ref{tab:hyper_wmpd}} summarizes all the hyperparameters used for the unlearning tasks on WMDP.
\begin{table}
    \centering
    \resizebox{0.5\textwidth}{!}{
    \begin{tabular}{l|c|c|c|c|c|c}
        \toprule[1pt]
        \textbf{Method} & \textbf{$\eta$} & \textbf{$\beta$} & \textbf{$\gamma$} & \textbf{$\alpha$} & \textbf{$\lambda$} & $c$ \\
        \midrule
        RMU  & $5\times 10^{-5}$ & - & - & - & 1200 & 6.5 \\
        \midrule
        NPO  & $10^{-5}$ & 0.1  & -  & - & 1.0 & - \\
        \midrule
        SimNPO  & $10^{-5}$ & 0.1  & -  & 0.0 & 1.0 & - \\
        \midrule
        NGDiff--RMU & $10^{-5}$ & -  & -  & - & - & 6.5 \\
        \midrule
        BLUR--NPO  & $2\times 10^{-6}$ & 0.005  & 1.0  & - & - & - \\
        \midrule
        BLUR--RMU & $4\times 10^{-2}$ & -  & 0.00125  & - & - & 6.5\\
        \bottomrule
    \end{tabular}
    }
    \caption{Hyperparamters for various unlearning methods on WMDP benchmark}
    \label{tab:hyper_wmpd}
\end{table}

\subsection{Additional Experiment Results}
\begin{figure}[H]
    \centering
    \includegraphics[width=0.32\linewidth]{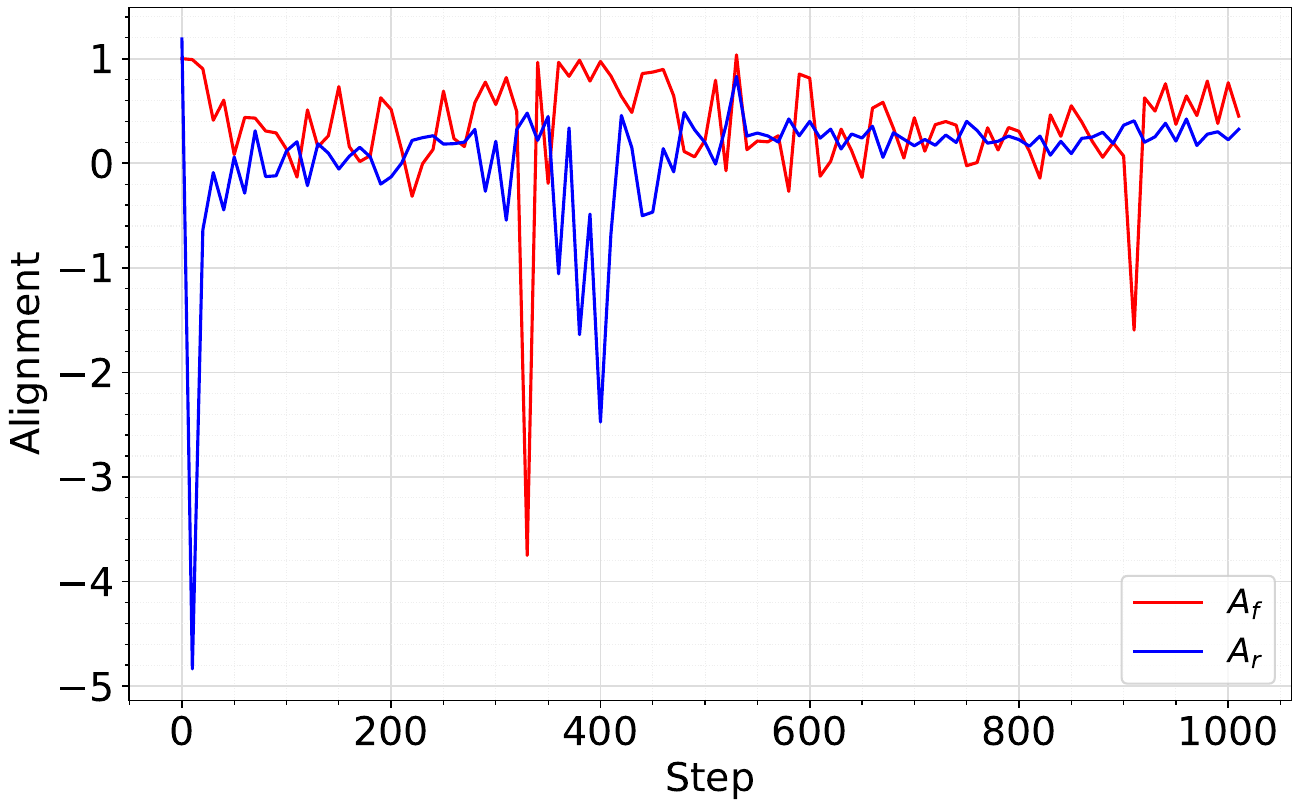}
    \includegraphics[width=0.32\linewidth]{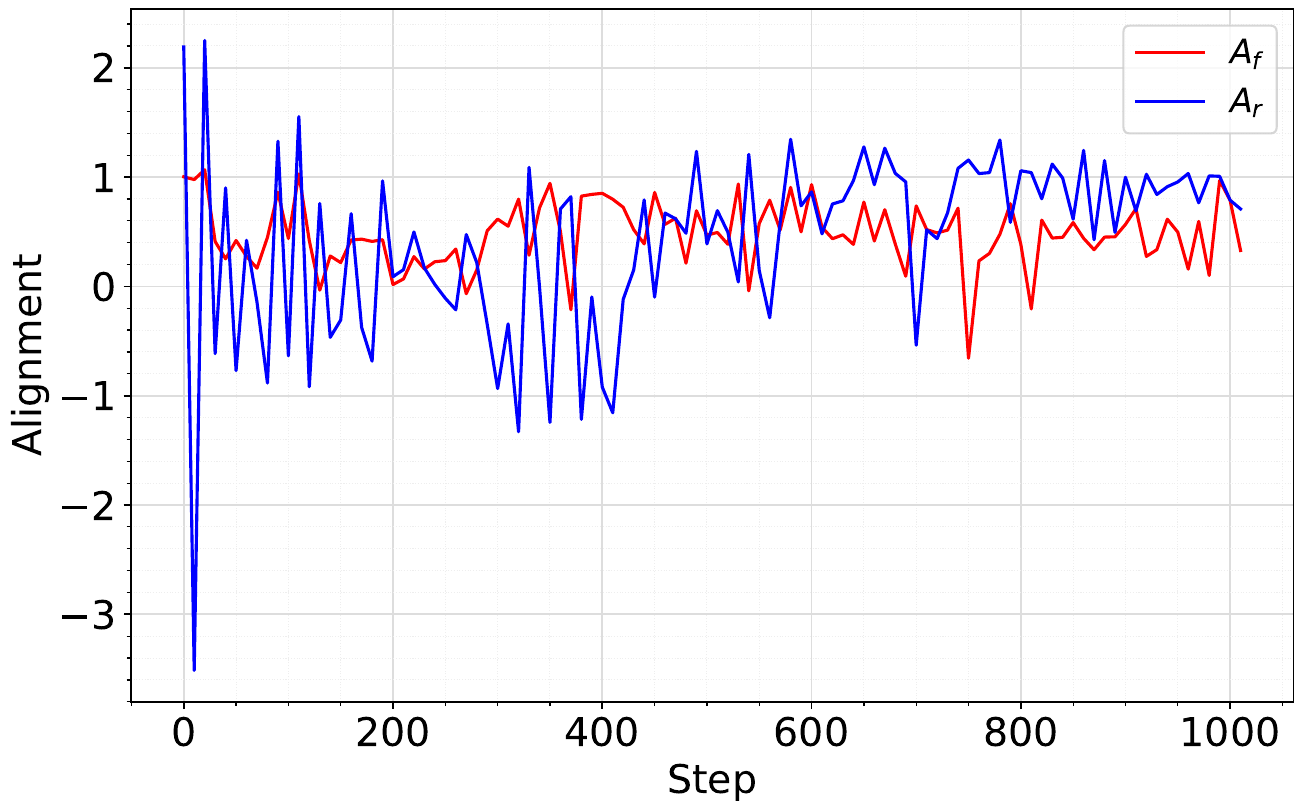}
    \includegraphics[width=0.32\linewidth]{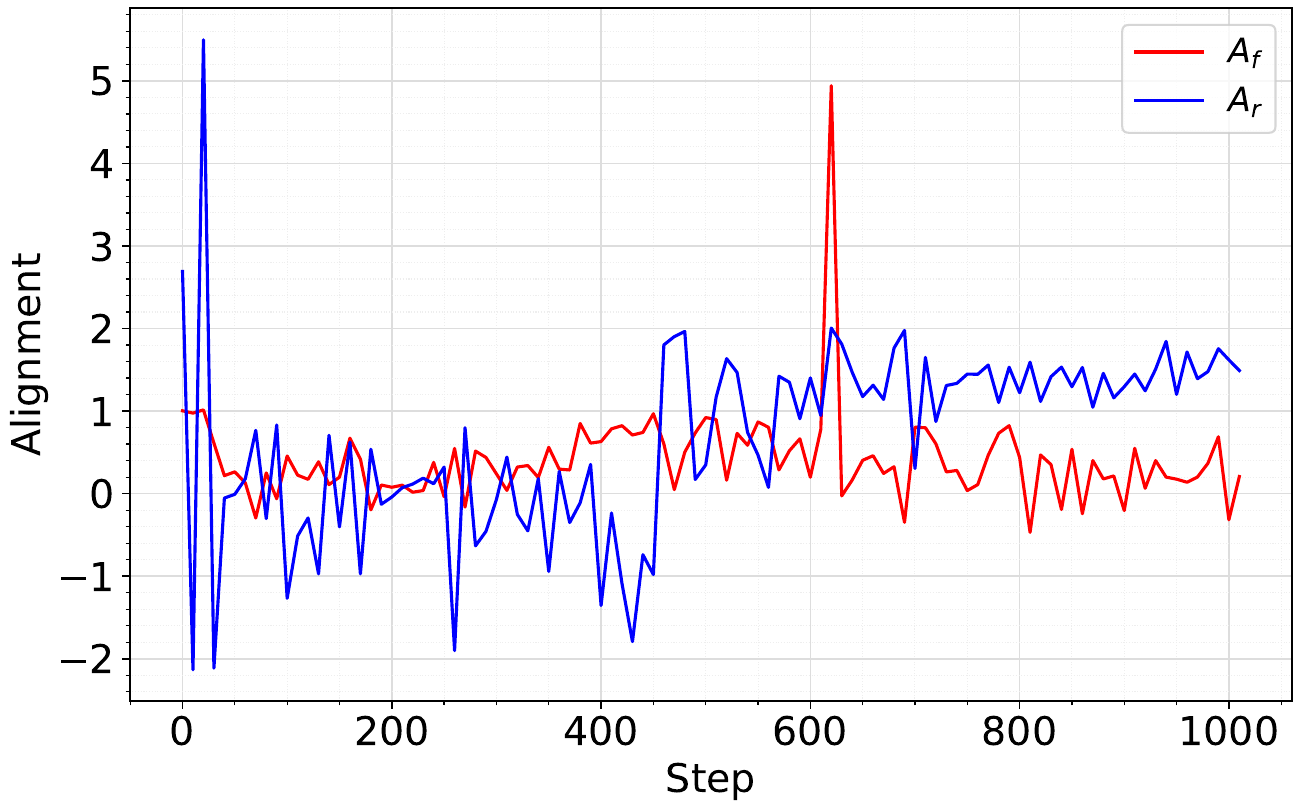}
    \caption{Alignment values of forget and retain objectives as defined in~\eqref{eq:alig} on MUSE-News dataset using LLaMa2-7B model vs. step with $\lambda=0.5$ (top-left), $\lambda=1.5$ (top-right), and $\lambda=2$ (bottom). }
    \label{fig:alig_extra}
\end{figure}

\begin{figure}[H]
    \centering
    \includegraphics[width=0.32\linewidth]{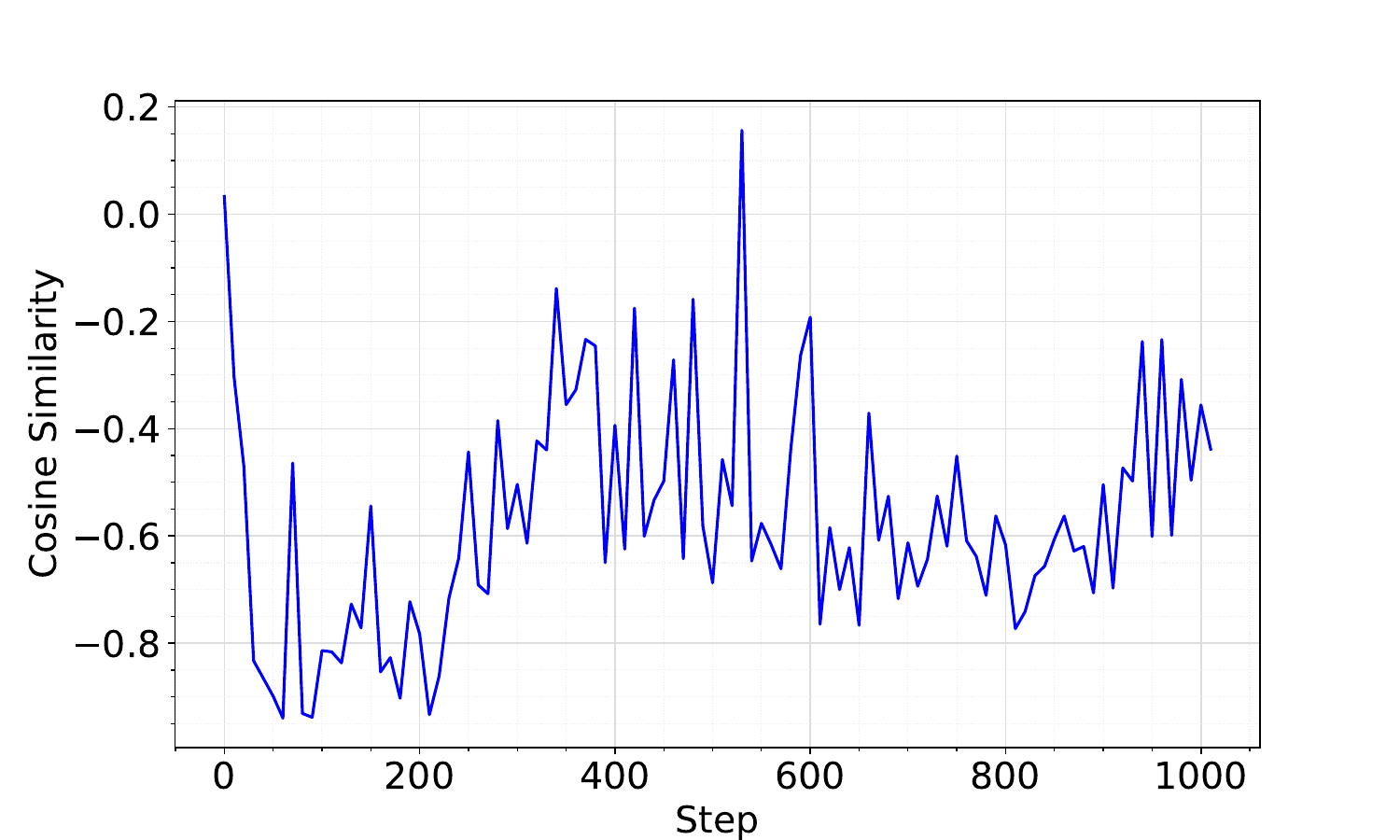}
    \includegraphics[width=0.32\linewidth]{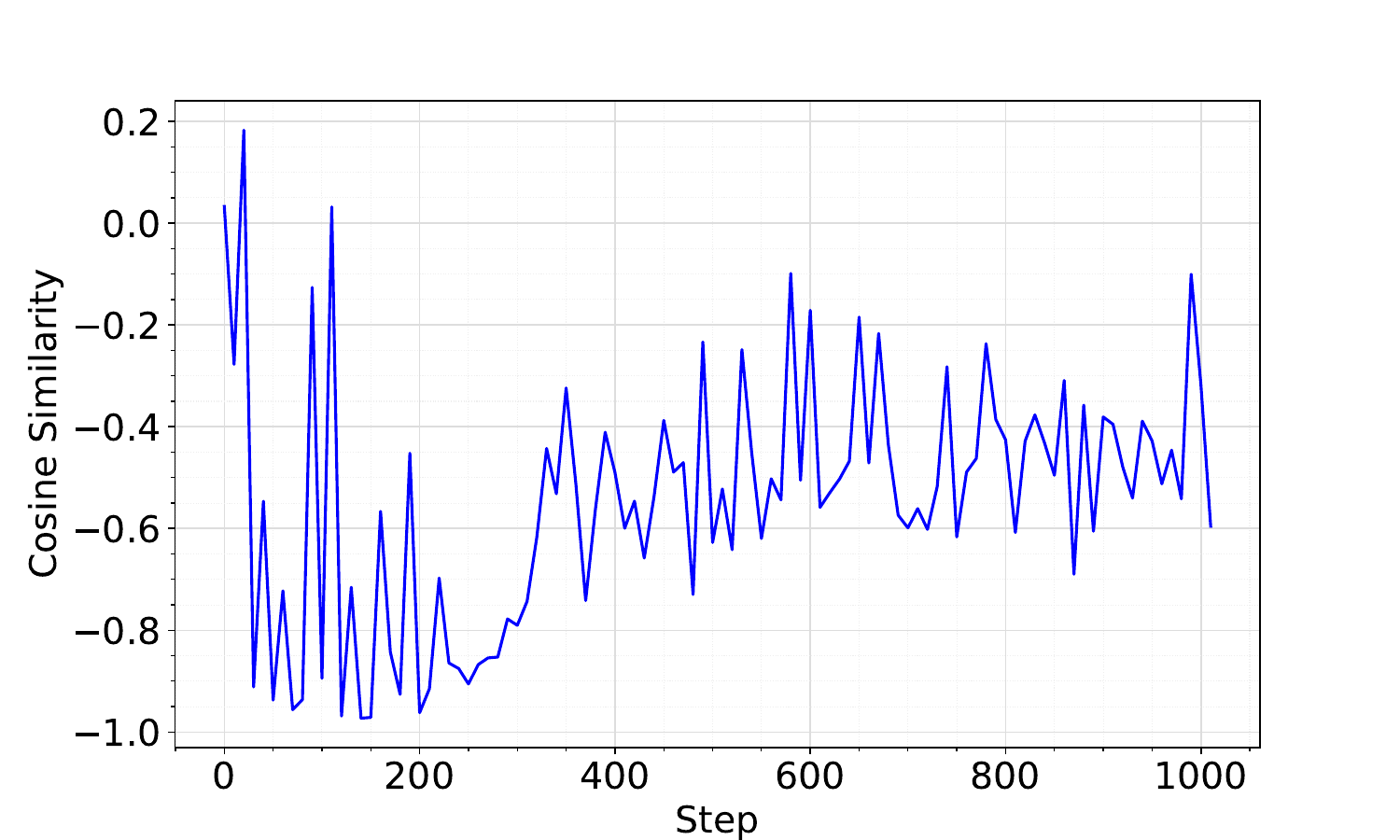}
    \includegraphics[width=0.32\linewidth]{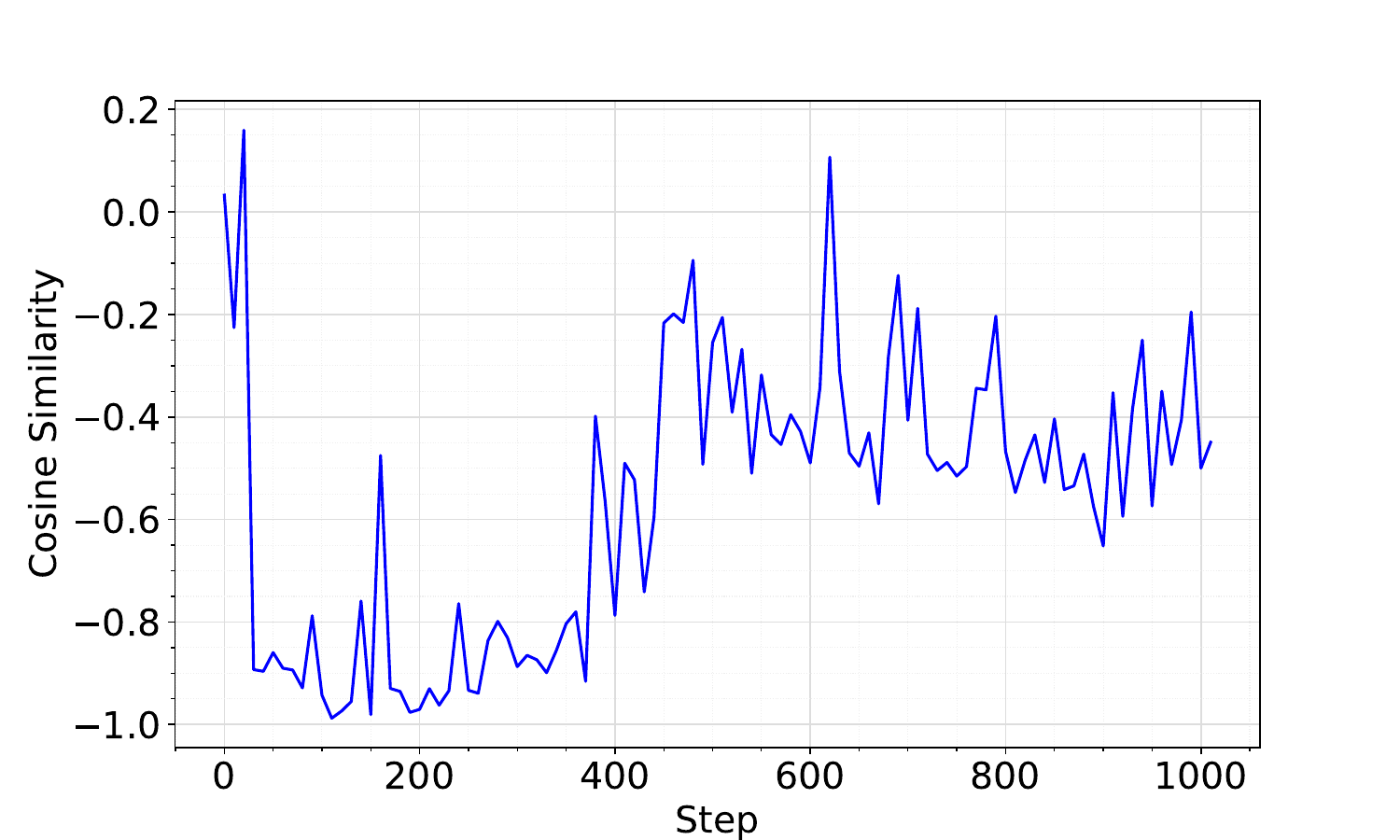}
    \caption{Cosine similarity of the gradient forget and retain loss functions using NPO method on MUSE-News dataset using LLaMa2-7B model vs. step with $\lambda=0.5$ (top-left), $\lambda=1.5$ (top-right), and $\lambda=2$ (bottom). }
    \label{fig:cos_extra}
\end{figure}

\begin{figure}[H]
    \centering
    \includegraphics[width=0.32\linewidth]{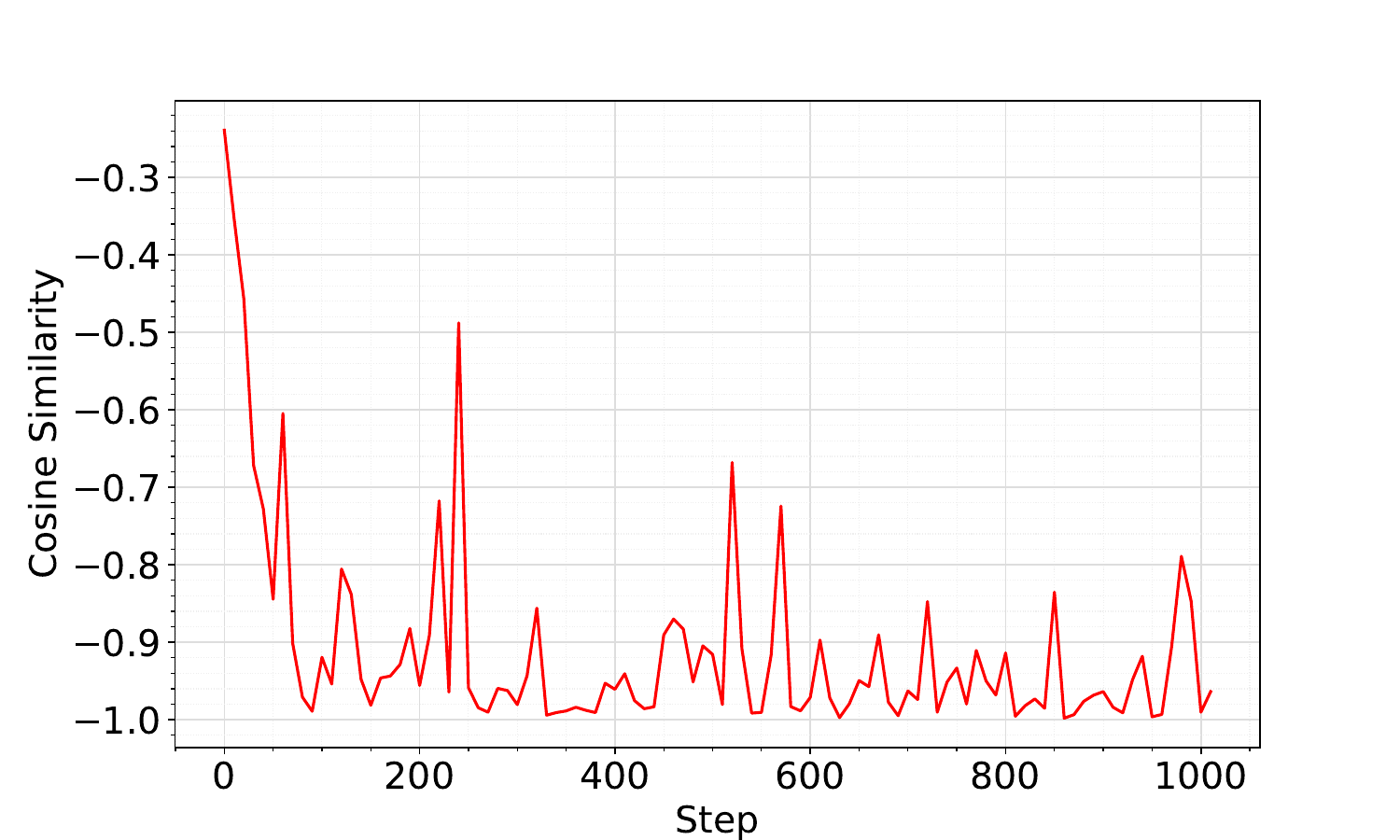}
    \includegraphics[width=0.32\linewidth]{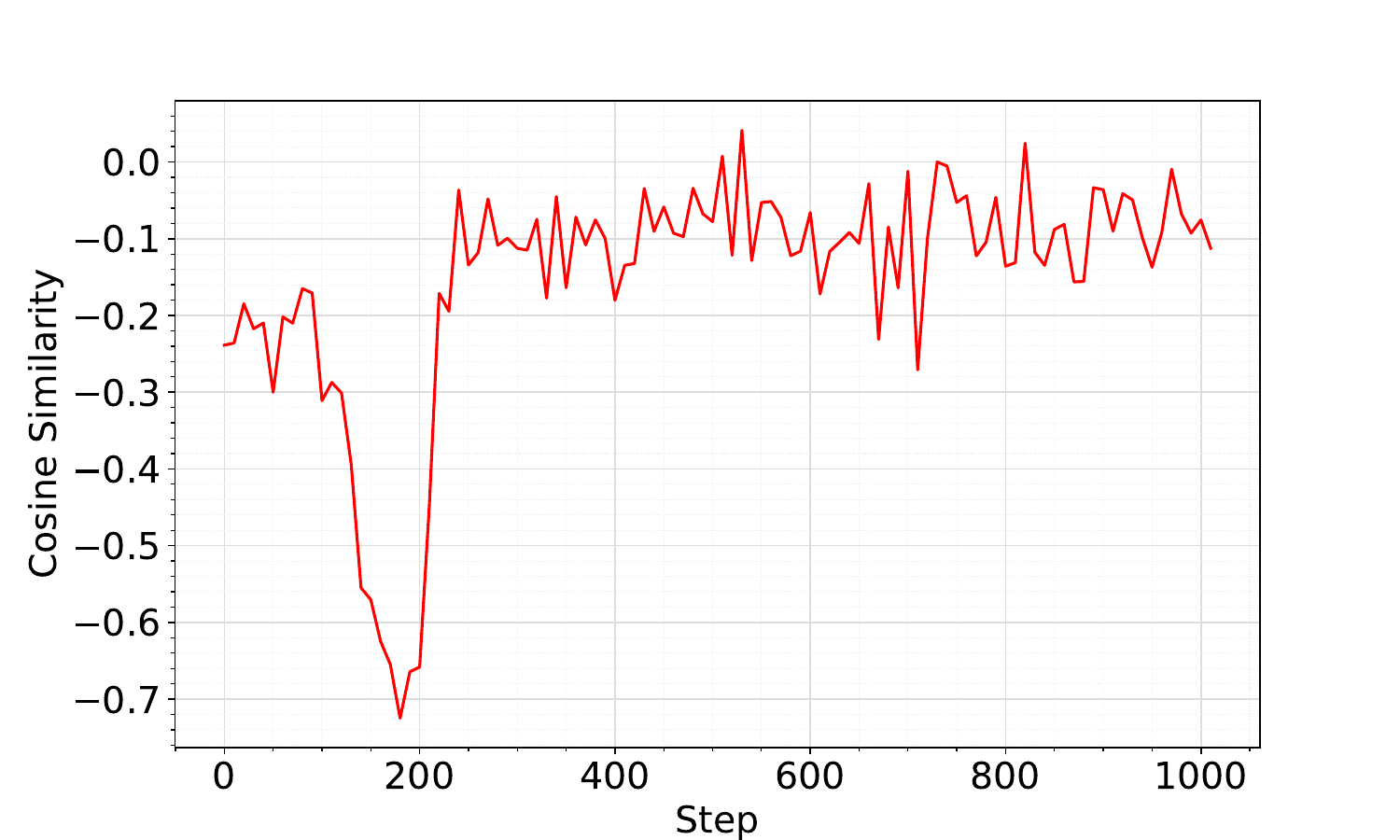}
    \includegraphics[width=0.32\linewidth]{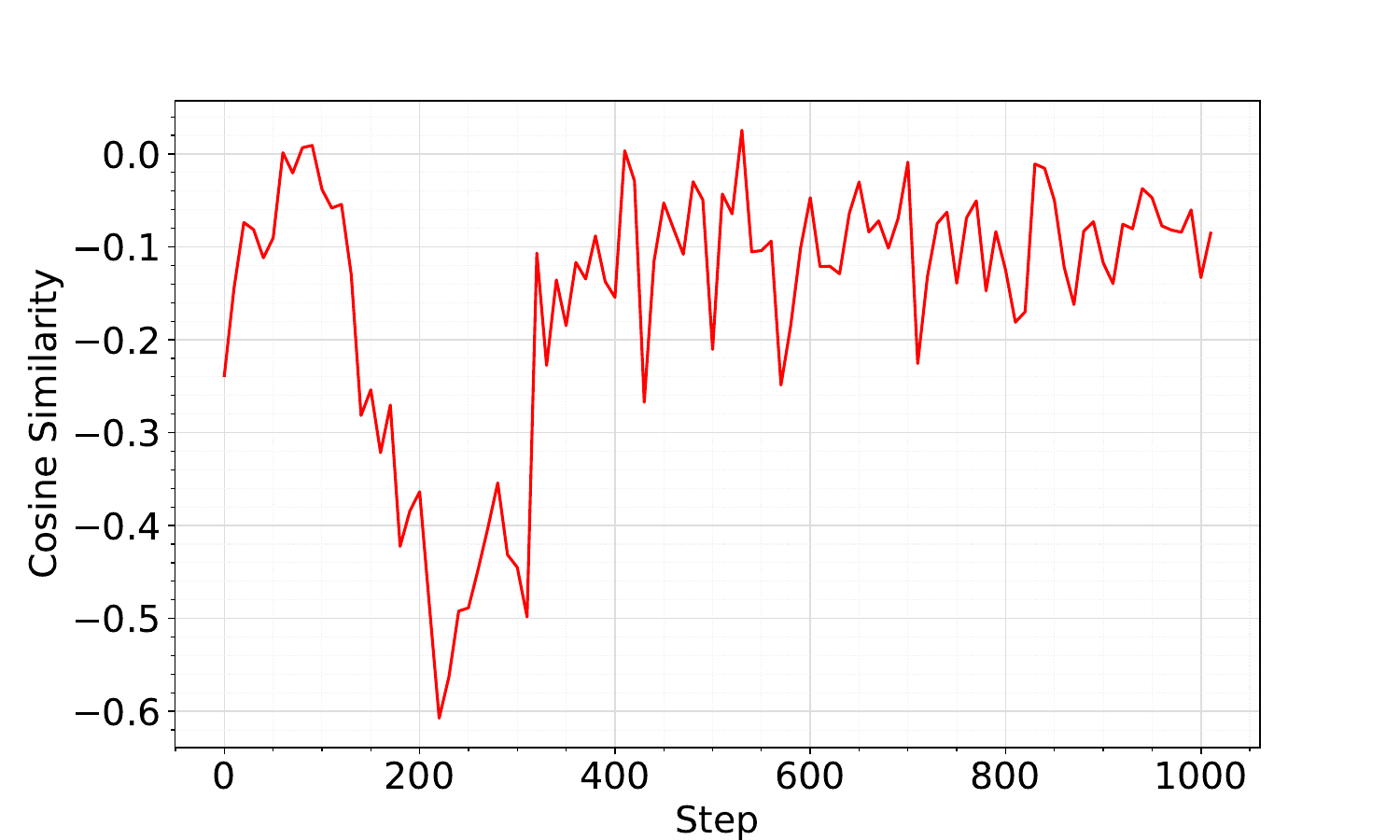}
    \caption{Cosine similarity of the gradient forget and retain loss functions using GradDiff method on MUSE-News dataset using LLaMa2-7B model vs. step with $\lambda=0.5$ (top-left), $\lambda=1$ (top-right), and $\lambda=1.5$ (bottom). }
    \label{fig:cos_extra_ga}
\end{figure}

Here, we discuss our experimental results further. As \textbf{Fig.~\ref{fig:alig_extra}} shows, the update direction $\hat{u}(\bft)$ cannot consistently prioritize the forget loss over the retain loss, even for various values of $\lambda$. This further corroborates that the regularized problem formulation in~\eqref{eq:reg_f} fails to achieve a proper balance between the forget and retain losses. \textbf{Fig.~\ref{fig:cos_extra}} and~\textbf{Fig.~\ref{fig:cos_extra_ga}} demonstrate that the retain and forget gradients conflict over the unlearning steps across various unlearning schemes and values of $\lambda$.

\begin{table}[H]
\begin{center}
\resizebox{0.52\textwidth}{!}{
\begin{tabular}{c|c}
\toprule[1pt]
\multicolumn{2}{c}{\begin{tabular}{c}
\textbf{Forget Set} \\
\footnotesize{\texttt{Who defended the plan for juryless rape trials}} \\ \footnotesize{\texttt{in Scotland by pointing to a 'weight of evidence'}} \\ \footnotesize{\texttt{that juries are affected by rape myths and misconceptions?}}\\
\end{tabular}} \\
\midrule
Answer & \begin{tabular}{c}
     \footnotesize{\texttt{\colorbox{red!20}{First Minister Humza Yousaf}}}
\end{tabular}\\
\midrule
GA & \begin{tabular}{c}
\footnotesize{\texttt{\colorbox{green!20}{\textbullet\textbullet\textbullet\textbullet\textbullet\textbullet\textbullet\textbullet\textbullet\textbullet\textbullet\textbullet\textbullet\textbullet\textbullet\textbullet\textbullet\textbullet\textbullet\textbullet\textbullet\textbullet\textbullet\textbullet\textbullet\textbullet\textbullet\textbullet\textbullet\textbullet\textbullet\textbullet}}} \\  
\end{tabular} \\
\midrule
GradDiff & \begin{tabular}{c}
     \footnotesize{\texttt{\colorbox{red!20}{1st Minister Humza Yousaf}}}\\ \footnotesize{\texttt{\colorbox{yellow!20}{\textbullet \ None 'I'm not a gambler'}}}\\
     \footnotesize{\texttt{\colorbox{yellow!20}{- Selby and the Crucible \textbullet \ None The}}} \\  
\end{tabular} \\
\midrule
NGDiff & \begin{tabular}{c}
     \footnotesize{\texttt{\colorbox{red!20}{1st Minister Humza Yousaf}}}
\end{tabular}\\
\midrule
NPO & \begin{tabular}{c}
     \footnotesize{\texttt{\colorbox{red!20}{1st Minister Humza Yousaf}}}
\end{tabular}\\
\midrule
SimNPO & \footnotesize{\texttt{\colorbox{red!20}{1st Minister Humza Yousaf}}}\\
\midrule
BLUR--NPO & \footnotesize{\texttt{\colorbox{green!20}{100\% of lawyers surveyed}}}\\
\midrule
\multicolumn{2}{c}{\begin{tabular}{c}
\textbf{Retain Set} \\
\footnotesize{\texttt{How many years ago did David and Janice Hunter retire to Cyprus?}}\\
\end{tabular}} \\
\midrule
Answer & \begin{tabular}{c}
     \footnotesize{\texttt{\colorbox{green!20}{20 years ago}}}
\end{tabular}\\
\midrule
GA & \begin{tabular}{c}
\footnotesize{\texttt{\colorbox{red!20}{\textbullet\textbullet\textbullet\textbullet\textbullet\textbullet\textbullet\textbullet\textbullet\textbullet\textbullet\textbullet\textbullet\textbullet\textbullet\textbullet\textbullet\textbullet\textbullet\textbullet\textbullet\textbullet\textbullet\textbullet\textbullet\textbullet\textbullet\textbullet\textbullet\textbullet\textbullet\textbullet}}} \\  
\end{tabular} \\
\midrule
GradDiff & \begin{tabular}{c}
     \footnotesize{\texttt{\colorbox{green!20}{20}}} \footnotesize{\texttt{\colorbox{yellow!20}{the 'Happy the the the the the' couple,}}}\\
     \footnotesize{\texttt{\colorbox{yellow!20}{as the 2023 'The Happy the the the' book and the}}} \\  
\end{tabular} \\
\midrule
NGDiff & \begin{tabular}{c}
     \footnotesize{\texttt{\colorbox{red!20}{25}}}\footnotesize{\texttt{\colorbox{green!20}{years}}}
\end{tabular}\\
\midrule
NPO & \begin{tabular}{c}
     \footnotesize{\texttt{\colorbox{red!20}{25}}}\footnotesize{\texttt{\colorbox{green!20}{years}}}
\end{tabular}\\
\midrule
SimNPO & \begin{tabular}{c}
    \footnotesize{\texttt{\colorbox{red!20}{25}}}\footnotesize{\texttt{\colorbox{green!20}{years}}}
\end{tabular}\\
\midrule
BLUR--NPO & \begin{tabular}{c}
\footnotesize{\texttt{\colorbox{green!20}{20 years ago}}}
\end{tabular}\\
\bottomrule[1pt]
\end{tabular}
} 
\caption{Examples of generated text from different unlearned models in the MUSE-News dataset. Failed unlearning is indicated by undesired answers highlighted in \colorbox{red!20}{red}, while successful unlearning is shown in \colorbox{green!20}{green} for desired responses. Repeated or irrelevant information is marked in \colorbox{yellow!20}{yellow}.} 
\label{tab:muse_news}
\vspace{-0.5cm}
\end{center}
\end{table}

\begin{figure}[h]
    \centering    \includegraphics[width=0.42\linewidth]{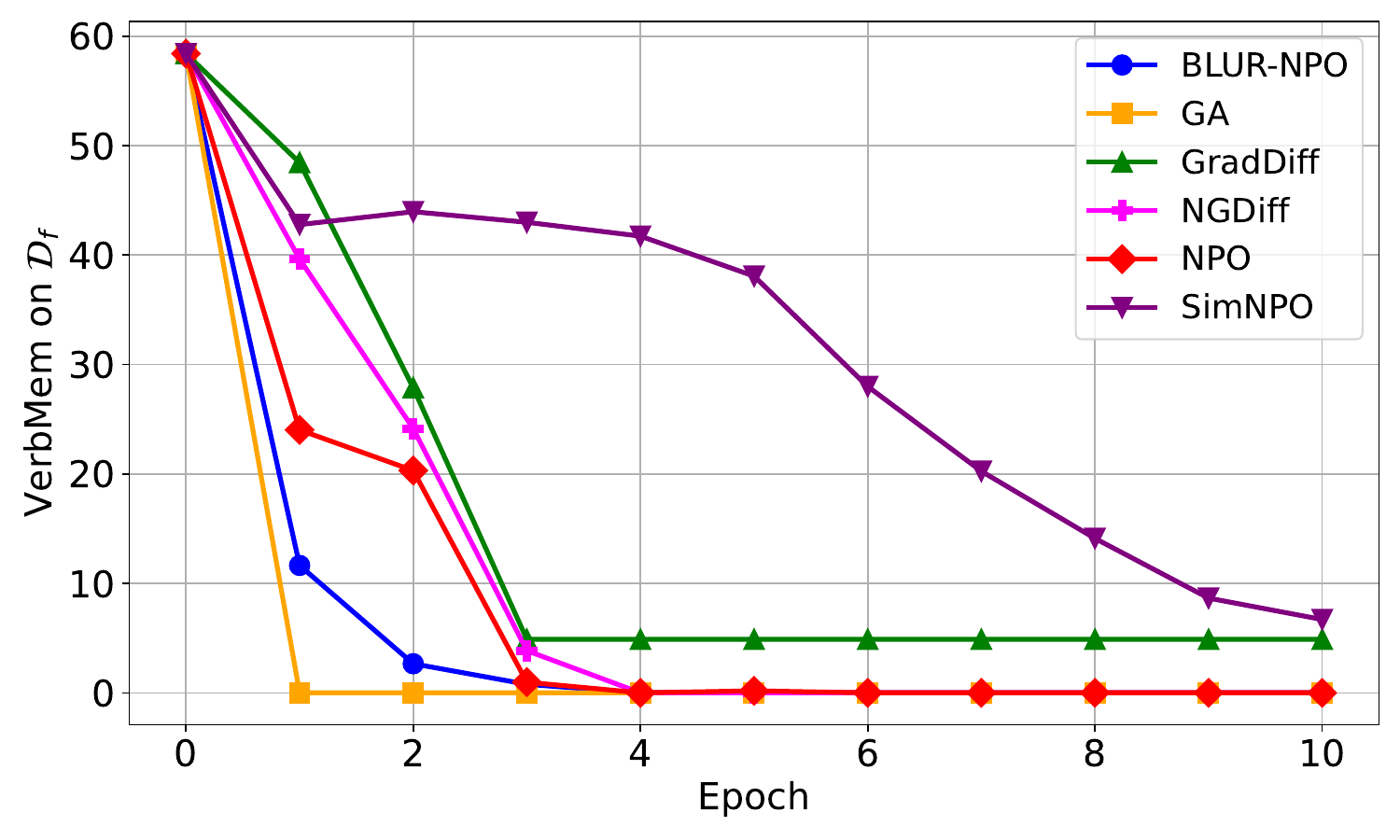}
    \includegraphics[width=0.42\linewidth]{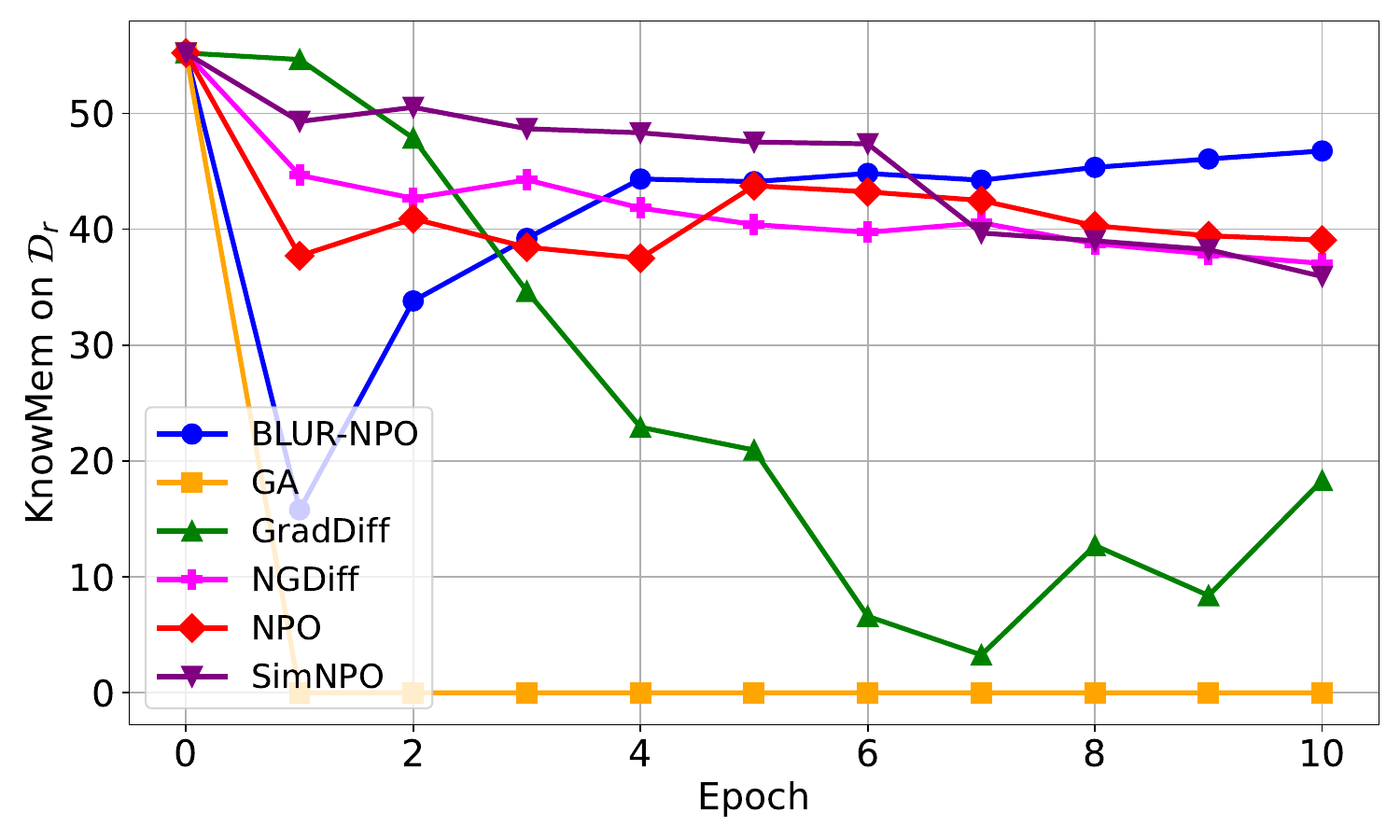}
    \vspace{-3mm}
    \caption{Verbatim memorization on the forget set $\mathcal{D}_f$ (top) and knowledge memorization on the retain set $\mathcal{D}_r$ (bottom) vs. optimization epochs, using various unlearning methods on the MUSE-News dataset.}
    \label{fig:verb_knw}
   \vspace{-0.4cm}
\end{figure}


\begin{figure}[H]
    \centering
    \begin{minipage}{0.48\linewidth}
        \centering
        \includegraphics[width=\linewidth]{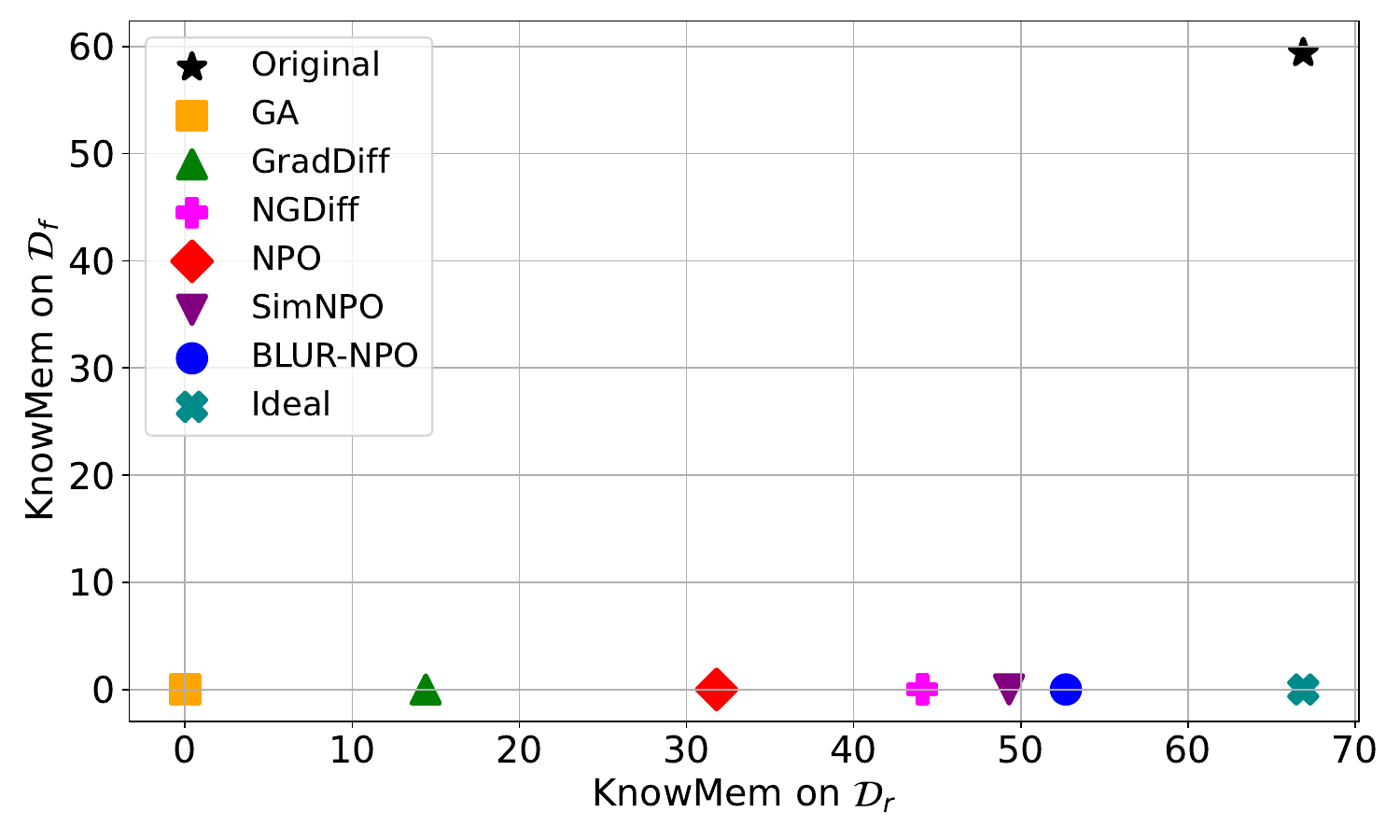}
        \caption{Trade-off between KnowMem values on the forget and retain datasets using different unlearning methods, LLaMA2-7B model, and MUSE-Books dataset where BLUR--NPO outperforms SOTA models.} 
        \label{fig:trade-off_books}
    \end{minipage}
    \hfill
    \begin{minipage}{0.48\linewidth}
        \centering
        \includegraphics[width=\linewidth]{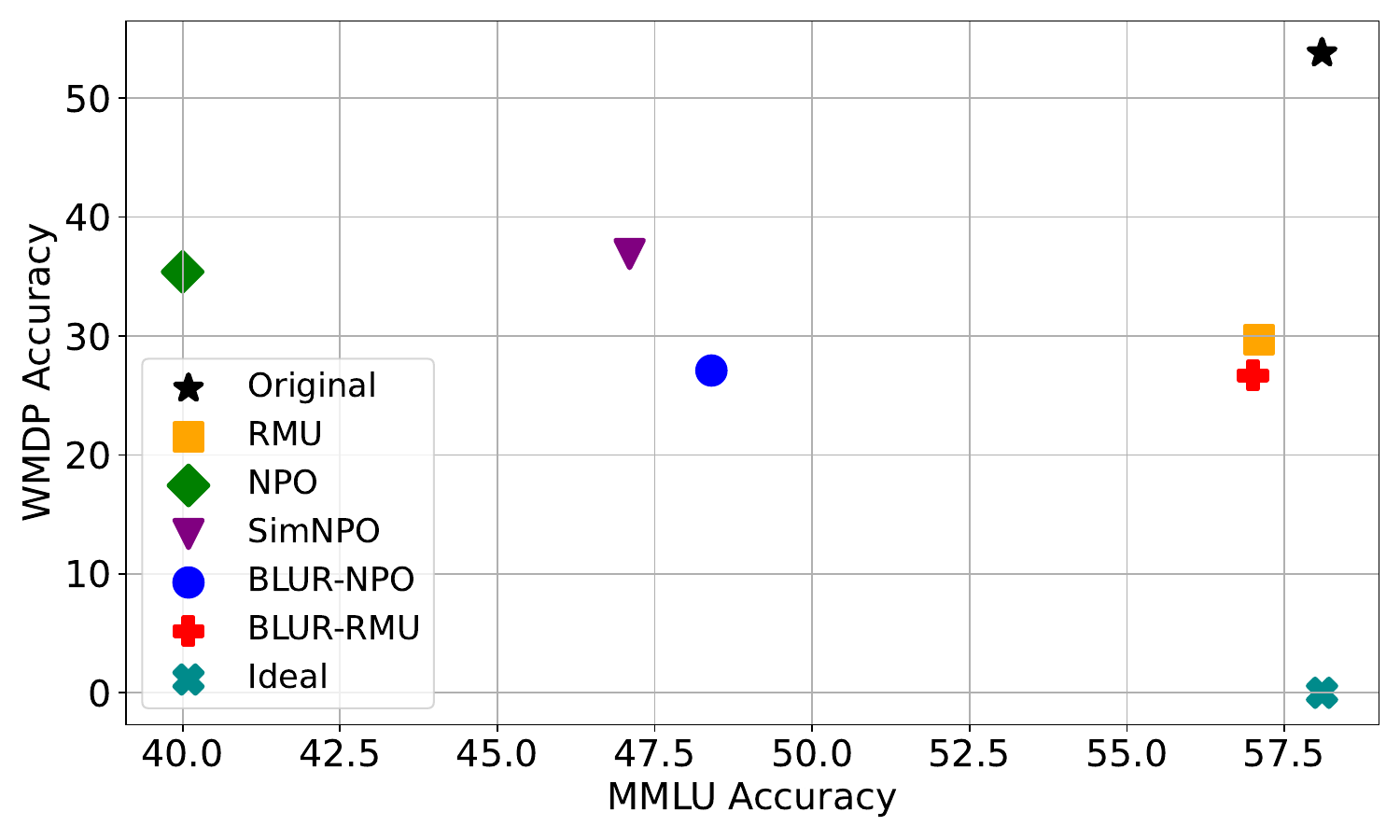}
        \caption{Trade-off between WMDP Accuracy and MMLU Accuracy using different unlearning methods with Zephyr-7B-beta model. Our algorithm BLUR--RMU outperforms RMU.}
        \label{fig:trade-off_wmdp}
    \end{minipage}
\end{figure}

\textbf{Fig.~\ref{fig:trade-off_books}} illustrates the KnowMem trade-off between the forget and retain datasets on MUSE-Books. As observed, all unlearning methods achieve no-KnowMem status on the forget set, while BLUR--NPO outperforms all other methods regarding retain KnowMem. \textbf{Fig.~\ref{fig:trade-off_wmdp}} presents the trade-off between model utility and unlearning efficiency on the WMDP dataset. In particular, BLUR--RMU outperforms the RMU method $10\%$ in WMDP accuracy while maintaining nearly the same performance in MMLU accuracy. 

\noindent{\textbf{Ablation Studies of BLUR--NPO on MUSE-News Corpus.}}

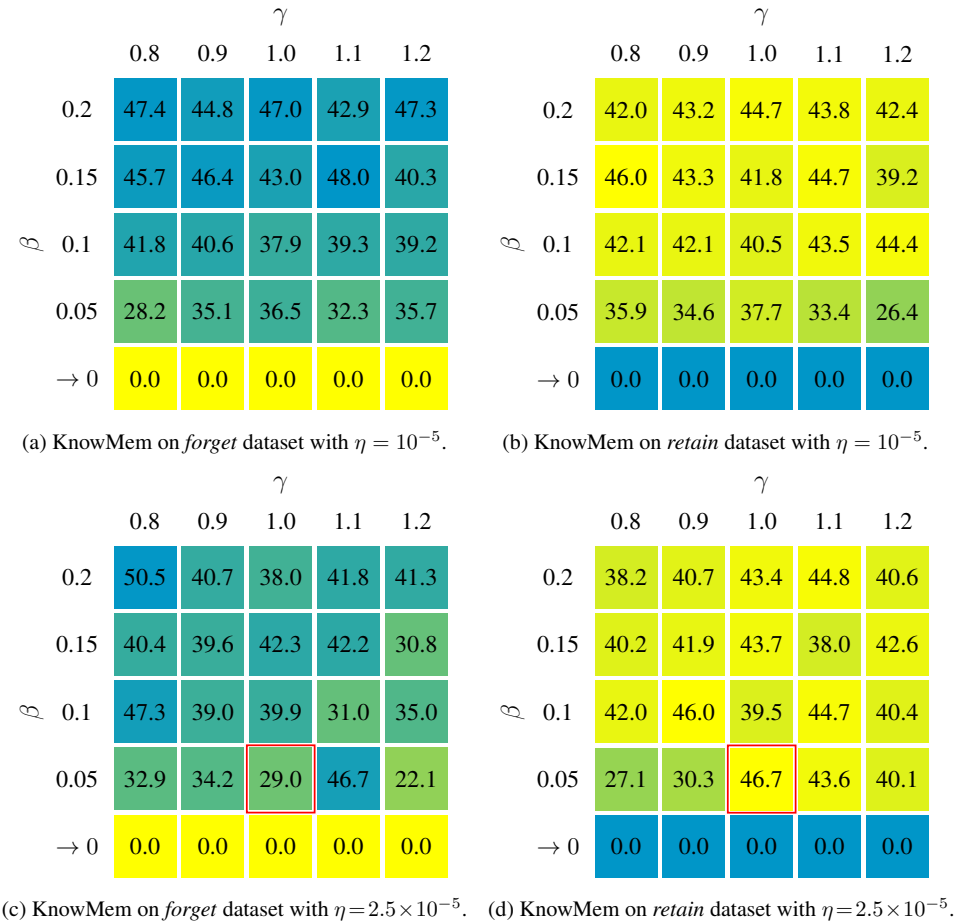
\begin{figure}[H]
    \centering
    \resizebox{0.8\textwidth}{!}{
    \begin{minipage}{0.5\textwidth}
        \centering
        \begin{tikzpicture}[scale=0.9]
            \pgfmathsetmacro{\vmin}{0.0}  
            \pgfmathsetmacro{\vmax}{48}
            \foreach \y [count=\n] in {
                {47.4,44.8,47.0,42.9,47.3},
                {45.7,46.4,43.0,48.0,40.3},
                {41.8,40.6,37.9,39.3,39.2},
                {28.2,35.1,36.5,32.3,35.7},
                {0.0,0.0,0.0,0.0,0.0},
            } {
                \foreach \x [count=\m] in \y {
                    \pgfmathsetmacro\colorval{100 * (\x - \vmin) / (\vmax - \vmin)}
                    \node[fill=bluegreen!\colorval! lightyellow, minimum size=10mm, text=black] at (\m*1.2,-\n*1.2) {\x};
                }
            }
            \foreach \a [count=\i] in {0.8, 0.9, 1.0, 1.1, 1.2} {
                \node[minimum size=10mm] at (1.2*\i, -0.2) {\a};
            }
            \foreach \a [count=\i] in {0.2, 0.15, 0.1, 0.05, $\rightarrow 0$} {
                \node[minimum size=8mm] at (0,-1.2*\i) {\a};
            }

            \node[rotate=0] at (3.6,0.45) {\large $\gamma$};
            \node[rotate=90] at (-0.8,-3.6) {\large $\beta$};
        \end{tikzpicture}
        \vspace{2mm}
        \par\small\text{(a) KnowMem on \textit{forget} dataset with $\eta = 10^{-5}$.}
    \end{minipage}
    \hfill
    \begin{minipage}{0.45\textwidth}
        \centering
        \begin{tikzpicture}[scale=0.9]
            \pgfmathsetmacro{\vmin}{0.0}  
            \pgfmathsetmacro{\vmax}{46}

            \foreach \y [count=\n] in {
                {42.0,43.2,44.7,43.8,42.4},
                {46.0,43.3,41.8,44.7,39.2},
                {42.1,42.1,40.5,43.5,44.4},
                {35.9,34.6,37.7,33.4,26.4},
                {0.0,0.0,0.0,0.0,0.0},
            } {
                \foreach \x [count=\m] in \y {
                    \pgfmathsetmacro\colorval{100 * (\vmax - \x) / (\vmax - \vmin)}
                    \node[fill=bluegreen!\colorval! lightyellow, minimum size=10mm, text=black] at (\m*1.2,-\n*1.2) {\x};
                }
            }
            \foreach \a [count=\i] in {0.8, 0.9, 1.0, 1.1, 1.2} {
                \node[minimum size=8mm] at (1.2*\i, -0.2) {\a};
            }
            \foreach \a [count=\i] in {0.2, 0.15, 0.1, 0.05, $\rightarrow 0$} {
                \node[minimum size=8mm] at (0,-1.2*\i) {\a};
            }

            \node[rotate=0] at (3.6,0.45) {\large $\gamma$};
            \node[rotate=90] at (-0.8,-3.6) {\large $\beta$};
        \end{tikzpicture} 
        \vspace{2mm}
        \par\small\text{(b) KnowMem on \textit{retain} dataset with $\eta=10^{-5}$.}
    \end{minipage}
    }
    \resizebox{0.8\textwidth}{!}{
    \begin{minipage}{0.5\textwidth}
    \vspace{2mm}
        \centering
        \begin{tikzpicture}[scale=0.9]
            \pgfmathsetmacro{\vmin}{0.0}  
            \pgfmathsetmacro{\vmax}{51}

            \foreach \y [count=\n] in {
                {50.5,40.7,38.0,41.8,41.3},
                {40.4,39.6,42.3,42.2,30.8},
                {47.3,39.0,39.9,31.0,35.0},
                {32.9,34.2,29.0,46.7,22.1},
                {0.0,0.0,0.0,0.0,0.0},
            } {
                \foreach \x [count=\m] in \y {
                    \pgfmathsetmacro\colorval{100 * (\x - \vmin) / (\vmax - \vmin)}
                    \node[fill=bluegreen!\colorval! lightyellow, minimum size=10mm, text=black] at (\m*1.2,-\n*1.2) {\x};
                }
            }
            \draw[red, thick] (3,-4.2) rectangle (4.2,-5.4);
            \foreach \a [count=\i] in {0.8, 0.9, 1.0, 1.1, 1.2} {
                \node[minimum size=8mm] at (1.2*\i, -0.2) {\a};
            }
            \foreach \a [count=\i] in {0.2, 0.15, 0.1, 0.05, $\rightarrow 0$} {
                \node[minimum size=8mm] at (0,-1.2*\i) {\a};
            }
            \node[rotate=0] at (3.6,0.45) {\large $\gamma$};
            \node[rotate=90] at (-0.8,-3.6) {\large $\beta$};
        \end{tikzpicture}
        \vspace{2mm}
        \par\small\text{(c) KnowMem on \textit{forget} dataset with $\eta \!=\! 2.5\!\times\! 10^{-5}$.}
    \end{minipage} 
    \hfill
    \begin{minipage}{0.45\textwidth}
    \vspace{2mm}
        \centering
        \begin{tikzpicture}[scale=0.9]
            \pgfmathsetmacro{\vmin}{0.0}  
            \pgfmathsetmacro{\vmax}{46}

            \foreach \y [count=\n] in {
                {38.2,40.7,43.4,44.8,40.6},
                {40.2,41.9,43.7,38.0,42.6},
                {42.0,46.0,39.5,44.7,40.4},
                {27.1,30.3,46.7,43.6,40.1},
                {0.0,0.0,0.0,0.0,0.0},
            } {
                \foreach \x [count=\m] in \y {
                    \pgfmathsetmacro\colorval{100 * (\vmax - \x) / (\vmax - \vmin)}
                    \node[fill=bluegreen!\colorval! lightyellow, minimum size=10mm, text=black] at (\m*1.2,-\n*1.2) {\x};
                }
            }
            \draw[red, thick] (3,-4.2) rectangle (4.2,-5.4);
            \foreach \a [count=\i] in {0.8, 0.9, 1.0, 1.1, 1.2} {
                \node[minimum size=8mm] at (1.2*\i, -0.2) {\a};
            }
            \foreach \a [count=\i] in {0.2, 0.15, 0.1, 0.05, $\rightarrow 0$} {
                \node[minimum size=8mm] at (0,-1.2*\i) {\a};
            }
            \node[rotate=0] at (3.6,0.45) {\large $\gamma$};
            \node[rotate=90] at (-0.8,-3.6) {\large $\beta$};
        \end{tikzpicture} 
        \vspace{2mm}
        \par\small\text{(d) KnowMem on \textit{retain} dataset with $\eta \!=\! 2.5\!\times\! 10^{-5}$.}
    \end{minipage}
    }
    \caption{KnowMem values on \textit{forget} and \textit{retain} datasets using BLUR--NPO unlearning method, LLaMA2-7B model, and MUSE-News corpus under two learning rates $\eta=10^{-5}$ and $\eta=2.5\times 10^{-5}$ with different combinations of hyperparameters of $\beta$ and $\gamma$.}
    \label{fig:ab}
\end{figure}

The proposed descent direction $u(\bft)$ in~\eqref{eq:u} exploits a hyperparameter $\gamma$ adjusting the amplitude of the forget gradient. Moreover, for BLUR--NPO, we use NPO loss for the forget objective, where the hyperparameter $\beta$ regulates the intensity of unlearning. As $\beta\rightarrow 0$, NPO loss converges to the GA loss. \textbf{Fig.~\ref{fig:ab}} illustrates the performance of KnowMem on the forget and retain datasets of BLUR--NPO for various values of $\beta$ and $\gamma$ on MUSE-News using the LLaMA2-7B model with two learning rates $\eta=10^{-5}$ and $\eta=2.5\times 10^{-5}$. As observed, a large value of $\beta$ fails to unlearn the forget set while maintaining good performance on the retain set. Moreover, the learning rate $\eta=10^{-5}$ is not large enough to unlearn the forget set while preserving the model utility, whereas a larger learning rate $\eta=2.5\times 10^{-5}$ with a relatively small $\beta=0.05$ and $\gamma=1.0$ achieves both objectives simultaneously. The unlearning metrics for this optimal setup are highlighted with red rectangles in \textbf{Fig.~\ref{fig:ab}}.

\end{document}